\documentclass[journal]{IEEEtran}
\usepackage{amsmath,amsfonts}
\usepackage{algorithmic}
\usepackage{array}
\usepackage[caption=false,font=normalsize,labelfont=sf,textfont=sf]{subfig}
\usepackage{textcomp}
\usepackage{stfloats}
\usepackage{url}
\usepackage{verbatim}
\usepackage{graphicx}
\usepackage{balance}

\usepackage{booktabs}
\usepackage{multirow}
\usepackage{color}

\begin{document}

\title{Dynamic Appearance Particle Neural Radiance Field}

\author{
Ancheng Lin*, Yusheng Xiang*, Jun Li, Mukesh Prasad

\thanks{Corresponding author: Jun Li.}

\thanks{Ancheng Lin, Jun Li, and Mukesh Prasad are with the School of Computer Science, Australian Artificial Intelligence Institute (AAII), University of Technology Sydney, Sydney, NSW 2007, Australia (e-mail: ancheng.lin@student.uts.edu.au; jun.li@uts.edu.au; mukesh.prasad@uts.edu.au).} 

\thanks{Yusheng Xiang is with the School of Automobile, Chang'an University, Xi'an 710000, China (e-mail: xiangyusheng@chd.edu.cn).} 
\thanks{* contributed equally to the work.}

}

\maketitle
\makeatletter
\def\ps@IEEEtitlepagestyle{
  \def\@oddfoot{\mycopyrightnotice}
  \def\@evenfoot{}
}
\def\mycopyrightnotice{
  {\footnotesize
  \begin{minipage}{\textwidth}
  \centering
  Accepted by IEEE Transactions on Circuits and Systems for Video Technology, with DOI: 10.1109/TCSVT.2025.3540792
  \\
  Copyright~\copyright~2025 IEEE. Personal use of this material is permitted. However, permission to use this  \\ 
  material for any other purposes must be obtained from the IEEE by sending a request to pubs-permissions@ieee.org.
  \end{minipage}
  }
}

\newcommand{\ve}[1]{{\boldsymbol{#1}}}
\newcommand{\set}[1]{{\mathbf{#1}}}
\newcommand{\grid}[1]{{\mathbf{#1}}}
\newcommand{\modelname}[0]{DAP-NeRF}
\newcommand{\revised}[1]{{\textcolor{red}{#1}}}
\newcommand{\suggest}[1]{{\textcolor{blue}{#1}}}
\newcommand{\issue}[1]{{\textcolor{red}{#1}}}

\begin{abstract}
Neural Radiance Fields (NeRFs) have shown great potential in modeling 3D scenes. Dynamic NeRFs extend this model by capturing time-varying elements, typically using deformation fields. The existing dynamic NeRFs employ a similar Eulerian representation for both light radiance and deformation fields. This leads to a close coupling of appearance and motion and lacks a physical interpretation. In this work, we propose Dynamic Appearance Particle Neural Radiance Field (DAP-NeRF), which introduces particle-based representation to model the motions of visual elements in a dynamic 3D scene. DAP-NeRF consists of the superposition of a static field and a dynamic field. The dynamic field is quantized as a collection of {\em appearance particles}, which carries the visual information of a small dynamic element in the scene and is equipped with a motion model. All components, including the static field, the visual features and the motion models of particles, are learned from monocular videos without any prior geometric knowledge of the scene. We develop an efficient computational framework for the particle-based model. We also construct a new dataset to evaluate motion modeling. Experimental results show that DAP-NeRF is an effective technique to capture not only the appearance but also the physically meaningful motions in a 3D dynamic scene. Code is available at: https://github.com/Cenbylin/DAP-NeRF.

\end{abstract}


\IEEEpeerreviewmaketitle

\section{Introduction}
\begin{figure*}
    \centering
    \includegraphics[width=1.0\textwidth]{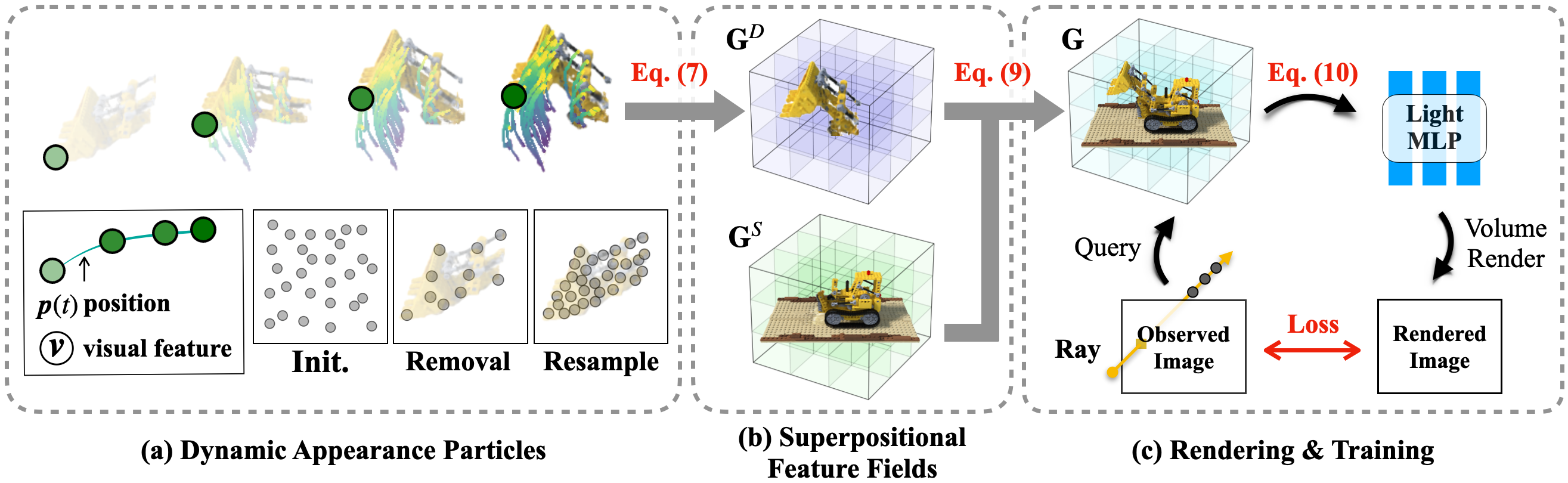}
    \caption{Method Overview. 
            \textbf{(a)} Particles represent observed movements.
            Each particle $\boldsymbol p$ corresponds to a small volume of material that represents a semantically meaningful part of a moving object. The position $\boldsymbol p(t)$ forms an explicit dynamic model of the object part. $\boldsymbol v$ denotes the visual feature of the particle (See Sec.~\ref{subs:particle-dyn-model}).
            \textbf{(b)} Particles are integrated into a grid-based feature field, enabling efficient computation and superposition with the static feature field (See Sec.~\ref{sec:eff_compute}).
            \textbf{(c)} Colors of rays are computed by querying sampled points on the superpositional radiance field and performing volume rendering. Photometric loss is then calculated to optimize the model (see Sec.~\ref{sec:optim}).
            \label{fig:overview}}
\end{figure*}

\IEEEPARstart{R}{ecent} advancements in neural radiance fields (NeRFs) \cite{Mildenhall20} have shown remarkable success in modeling 3D scenes with a continuous representation. These models facilitate high-fidelity rendering from novel viewpoints without requiring explicit geometry. One of the most promising extensions is Dynamic NeRFs \cite{Pumarola21_D-NeRF, Liu22_DeVRF}, which aims to model dynamic scenes by incorporating a time-varying radiance field. There are two primary schemes: directly adding a time dimension or introducing a deformation field before the canonical radiance field. The deformation-field scheme separately represents the motion and appearance of a dynamic scene and has gained popularity due to its effectiveness.


Existing models employ an Eulerian formulation for both the deformation and appearance fields. A learnable field model, such as a group of neural networks, uses the 3D Euclidean coordinates as its input query and outputs the desired physical quantities (field variables). This scheme is effective when the field needs to be defined throughout the entire Euclidean space, such as when considering the light scattering characteristics over a three-dimensional scene. However, the Eulerian field representation can be problematic when the quantities of interest are confined to specific regions or supported on a sub-manifold within the Euclidean space, for instance, within a solid object or on the surface. It might waste model capacity and make the subsequent use of the model inconvenient. For example, existing methods that apply dynamic NeRF to explore the interaction between objects and the environment require manual preprocessing like shape extraction \cite{qiao2022neuphysics} and inverse deformation estimation \cite{Chen22_VEO}.

To address these limitations, we propose to employ particle-based representation for moving or deforming objects within dynamic scenes. In our method, particles represent a finite approximation of the distribution of physical quantities that determine appearance. More specifically, in a volumetric rendering scheme, if the light properties, color, and scattering probability at a location $\ve x \in \mathcal{X}$ are determined by a view-independent physical feature $\ve f$ and the viewpoint, then the spatial distribution of $\ve f$ over $\mathcal{X}$ becomes the primary focus of modeling. For this purpose, particles are used to quantize the space $\mathcal{X}$, with the advantage that the distribution of $\ve f$ can be made time-varying by using a movement model for the particles. This particle-based representation can be integrated with an existing Eulerian appearance field that represents static elements. Therefore, we achieve a hybrid (static-Eulerian, dynamic-Lagrangian) NeRF model that can learn from monocular videos using only photometric supervision. 

In summary, this work introduces the Dynamic Appearance Particle Neural Radiance Field (\textbf{DAP-NeRF}). The major contributions are as follows:

\begin{itemize}
\item We have designed a hybrid framework of radiance field models. The dynamic elements of the fields are described using a particle-based representation, which corresponds to a Lagrangian approach to field models. The introduction of the Lagrangian model complements the widely adopted Eulerian dynamic NeRFs, which specify the static elements of the scene in our framework. The hybrid framework serves as more than an adequate appearance model of dynamic scenes. The employed particles also provide an explicitly interpretable and physically meaningful description of the motions.

\item We have developed an efficient and effective computational structure for the proposed hybrid framework. This structure has the capability to automatically identify dynamic and static elements.

\item We have constructed a dataset and introduced a metric to evaluate the motion modeling of dynamic NeRFs.

\end{itemize}

Empirical studies have shown that our framework achieves state-of-the-art performance in novel view synthesis tasks. We've demonstrated that our framework produces particles that effectively capture the dynamics of moving objects, facilitating scene decoupling and motion editing. Moreover, the learned particle motion model shows superior quantitative results in motion modeling, surpassing existing methods that deform a canonical field.

\section{Related Work} \label{sec:review}
\subsection{Dynamic NeRF}
Neural Radiance Field (NeRF) is a powerful technique in the field of learning 3D scene representations from a set of images \cite{Mildenhall20, TCSVT_nerfpointlight, TCSVT_nerfdeblurr, TCSVT_texturedMesh}. Due to its rendering quality and flexibility, NeRF has been extended to various areas, including generative models \cite{Jiang2023_AAAI_generative} and SLAM \cite{Li23_DIM-SLAM}.

There has been a growing interest in developing dynamic NeRFs for scenes where objects are moving or undergoing deformation \cite{TCSVT_deformscene, TCSVT_dynamicgarment}. Advancements in this field successfully model dynamic scenes from synchronized multi-view videos \cite{Liu22_DeVRF, Lombardi19_NeuralVolumes, Wang22_FourierPlenOctrees} and even monocular videos \cite{Pumarola21_D-NeRF, Tretsch21_NR-NeRF, Fang22_TINeuVox}. 

The first category of methods learns a time-conditional radiance field by adding a temporal input to the static radiance network or constructing a 4D volume \cite{Gao21_tnerf1, Xian21_tnerf2, Li21_tnerf3, Gan2022-V4D}. Methods that employ a time-conditional neural network produce completely different radiance and density fields at different time steps, which may lead to a severely under-constrained problem. These methods typically require additional supervision or regularization, such as depth and optical flow (as in Video-NeRF \cite{video_nerf} and NSFF \cite{NSFF}), or by incorporating multi-view observations \cite{DyNeRF}. Recently, the factorization of 4D volumes \cite{Keil_K-Planes} has yielded more efficient and less redundant representations, enabling more efficient training.

Another series of methods, known as deformable NeRFs \cite{Pumarola21_D-NeRF, Liu22_DeVRF, Tretsch21_NR-NeRF, Fang22_TINeuVox, Guo22_NDVG, ParkSBBGSM21_Nerfies}, decouple the model into a deformation field and a static canonical radiance field. The most commonly used technique in this category is backward deformation, which maps current 3D points back to a canonical space. While straightforward and efficient, this technique can struggle with motion inconsistencies in topologically complex scenes. Subsequent works such as \cite{Park21_HyperNeRF, song2022nerfplayer} have sought to address these challenges. On the other hand, \cite{ForwardFlowDNeRF} recognizes the discontinuity issues of the backward deformation and proposes using forward deformation instead. Nonetheless, challenges such as mapping ambiguity \cite{Uzolas23_Articulated} in the deformation field still persist, especially when modeling empty space. Building upon this background, this paper focuses on resolving issues induced by the Eulerian deformation field.


\subsection{Efficient Distributed Representation}
The original NeRF \cite{Mildenhall20} mapping raw spatial coordinates to light radiance can be time-consuming due to a large number of MLP forward passes. A recent advance is to distribute the field representation to individual voxels. DVGO \cite{Sun22_DVGO} employs a dense voxel grid and a small MLP to significantly accelerate NeRF training and rendering. Instant-NGP \cite{mueller2022instant} uses a hashing technique to reduce voxel grid storage costs and improve optimization speed and efficiency. 

Dynamic NeRFs have also adopted the representations introduced to static NeRFs. \cite{Fang22_TINeuVox} are the first to use voxel grids in the canonical radiance field, showing high training efficiency. \cite{Guo22_NDVG} improve the deformation field with a hybrid representation (voxel grid and MLP). \cite{Park23_hash4d} and \cite{Cao2023HexPlane} propose using 4D hash grids to decrease memory usage. \cite{Keil_K-Planes} introduce a factorization approach to the 4D grid, achieving a decomposition into static and dynamic components used in NeRFs.

\subsection{Point-based Representation}
This sub-area aims to discretize a radiance field using 3D points. Point-NeRF \cite{Xu22_point-nerf} was an early attempt using a point cloud representation encoding local geometry and appearance. However, it is restricted to static scenes and requires depth data. NeuroFluid \cite{Guan22_NeuroFluid} integrates particles into NeRF for fluid modeling, but lacks realistic texture and only works on simulated scenes. 
The very recent 3D Gaussian Splatting (3DGS) model \cite{3DGS} utilizes a purely explicit representation and employs fast point rasterization based on $\alpha$-blending, as opposed to the volumetric ray tracing used in NeRF methods. There are several works that extend 3DGS to dynamic scenes, such as DeformableGS \cite{DeformableGS} and 4D-GS \cite{4DGS}.

Two concurrent works \cite{Li23_PAC-NeRF, Uzolas23_Articulated} focus on scenes that conform to specific physical models. They only address foreground dynamic objects with pre-extracted initial shapes. In contrast, our method employs the particle-based representation to capture and model scene dynamics without prior geometric knowledge.

More closely related to the presented work, ParticleNeRF \cite{Chakra22_ParticleNeRF} encodes a scene entirely using particles, where the particles represent a single moment of a 3D scene and can be incrementally adapted to the next moment. Our method fundamentally differs from ParticleNeRF in several key aspects: First, ParticleNeRF captures only a single momentary state of the scene, optimizing from one state to the next. This remains a static NeRF training process, where the scene is observed from multiple views. Our method, however, uses a particle motion network to continuously model scene dynamics across all time frames. Second, ParticleNeRF encodes the entire scene with particles and uses radius nearest neighbor searching to query appearance features. In contrast, our method applies appearance particles only to dynamic elements and constructs a grid-based field that integrates these particles with a static feature grid, enabling more efficient querying through linear interpolation. Third, ParticleNeRF demands costly data collection, e.g., scenes with 20 camera views per time frame. On the other hand, our method is designed to handle monocular video data with just one camera view at each time frame.

\newcommand{\x}{\ve{x}}
\newcommand{\denf}[1]{\textrm{MLP}_{\sigma}(#1)}
\newcommand{\clrf}[1]{\textrm{MLP}_{\ve{c}}(#1)}

\newcommand{\superwei}[2]{{w_{#2}({#1})}}
\newcommand{\staV}[1]{V^{\mathrm{s}}(#1)}
\newcommand{\dynV}[2]{{V^{\mathrm{d}}_{#2}(#1)}} 
\newcommand{\dynVField}[1]{{V^{\mathrm{d}}_{#1}}} 

\section{Dynamic Appearance Particle NeRF\label{sec:model}}
\begin{figure}
    \centering
    \includegraphics[width=0.45\columnwidth]{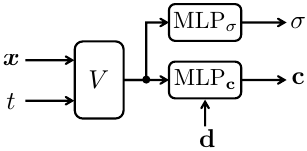}
    \caption{\label{fig:common_dyn_nerf} Commonly adopted structure of dynamic NeRFs.}
\end{figure}
\begin{figure*}
    \centering
    \includegraphics[width=0.8\textwidth]{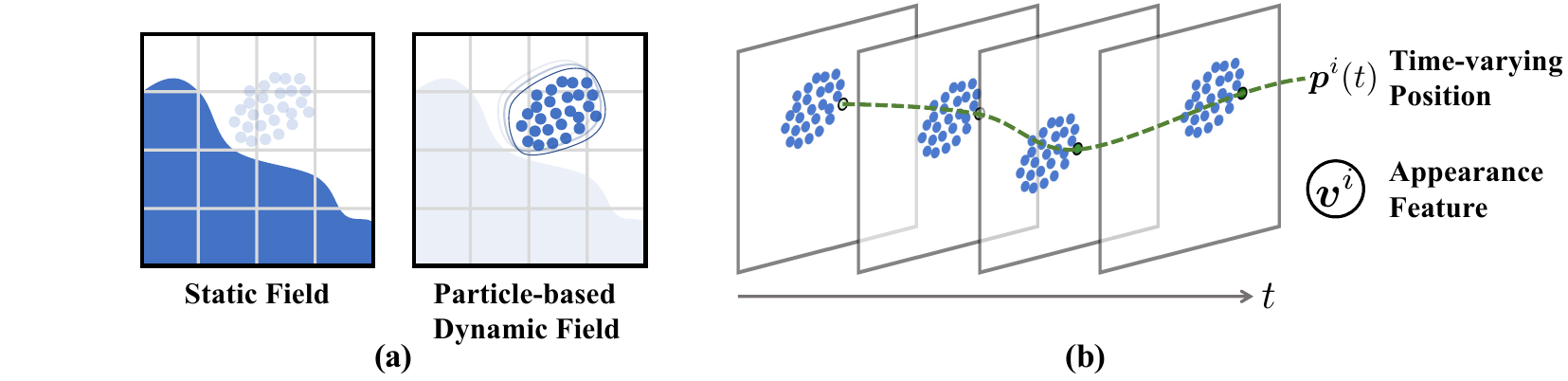}
    \caption{\label{fig:hybrid_rep} Overview of the proposed hybrid representation. \textbf{(a)} Superposition of dynamic and static field modeled by particle-based and Eulerian (voxel-grid-based) representations. \textbf{(b)} Two main attributes represented by particles. The particle trajectory $\ve{p}^i(t)$ is time-varying, while the appearance feature $\ve{v}^i$ is time invariant.}
\end{figure*}

Consider the dynamic radiance field model: A pixel of the camera frame at time $t$ is computed by casting a ray $\mathbf r(s)=\mathbf o + s\mathbf d$, $s \in \mathbb{R}^+$ from the camera center $\mathbf o$ to the pixel on the image plane:
\begin{align}
\mathbf C(\mathbf r, t) = \int_{s_n}^{s_f} T(s) \sigma(\mathbf r(s), t)\mathbf{c}(\mathbf r(s), \mathbf d, t)ds
\end{align}
where $\mathbf C(\mathbf r, t)$ denotes the pixel color rendered at time $t\in \mathbb R^+$, $s_n$, $s_f$ denote the near and far bounds of the ray. $\mathbf r(\cdot)$ returns a 3D location along the ray. $T(s)=\exp\big(-\int_{s_n}^{s}\sigma(\mathbf r(s'), t)ds'\big)$ is the accumulated transmittance. Given a location and time $t$ (and direction $\mathbf d$), $\sigma(\cdot)$ stands for the volume density, and $\mathbf c(\cdot)$ stands for the emitted radiance.

An effective strategy to specify the model of $\sigma(\cdot)$ and $\mathbf{c}(\cdot)$ is to learn small MLPs and localized feature vectors distributed at grid nodes in the 3D region of interest \cite{Fang22_TINeuVox}. Fig.~\ref{fig:common_dyn_nerf} displays the computational structure. 
The generic formulations of the components are as follows:
\begin{equation}
\begin{aligned}
     V(\ve x, t) &: \mathbb{R}^3 \times \mathbb{R} \mapsto \mathbb{R}^C \\
    \denf{\cdot} &: \mathbb{R}^C \mapsto \mathbb{R} \\
    \clrf{\cdot} &: \mathbb{R}^C \times \mathbb{R}^3 \mapsto \mathbb{R}^3
    \label{eq:feature_V_xt}
\end{aligned}
\end{equation}
where the majority of the field information is modeled by $V$. We adopt this framework and embed the Lagrangian dynamic model in $V$. However, it should be noted that the proposed technique is not tightly coupled to a specific overarching architecture. It is possible to integrate the proposed Lagrangian representation into alternative frameworks, such as Instant-NGP \cite{mueller2022instant} mentioned in Sec. \ref{sec:review}.



We model the feature field of the light radiance $V$ by the following superposition:
\begin{align}
V(\x, t) 
:= (1-\superwei{\x}{t}) \staV{\x}
+ \superwei{\x}{t} \dynV{\x}{t}  \label{eq:superimpose}
\end{align}
where $\staV{\x}$ and $\dynV{\x}{t}$ represent the static and dynamic components of the field, respectively. The static field component, $\staV{\x}$, employs a time-independent grid-based model. This aligns with the canonical feature field in \cite{Fang22_TINeuVox}. The particle-based dynamic field component, $\dynV{\x}{t}$, is proposed by this work and will be introduced in detail in the following subsections.
The superposition is determined by $\superwei{\x}{t}$, which takes value $1$ if the dynamic field is effective at $(\x, t)$ and $0$ otherwise. The support of the dynamic field (where it is effective) is implied by the particle representation of $\dynV{\x}{t}$. Fig. \ref{fig:hybrid_rep} (a) shows this superposition scheme.

It is helpful to notice a denotation convention due to the hierarchy of concepts that are adopted in this work. We denote time $t$ on the right-hand side of (\ref{eq:superimpose}) in subscripts, as opposed to an independent argument to the functions. This is to clarify the fact that the dynamic component $\dynV{\cdot}{\cdot}$ consists of an ensemble of particles. When considering the composition of particles making up $\dynV{\cdot}{t_0}$, the focus is on the discretization of a field at a specific time $t_0$. The time argument is fixed and does not affect the construction of the instantaneous status of $\dynV{\cdot}{t_0}$. The dynamics of the system are encoded in the individual particle models.


\subsection{Particle-based Dynamic Model\label{subs:particle-dyn-model}}
The dynamic field $\dynVField{t}$ is a continuous function, as a component of a NeRF model. We are concerned with integrating $\dynVField{t}$ in finite volumes. Therefore, we formulate $\dynVField{t}$ as
\begin{align}
\dynV{\ve x}{t} &= (\dynVField{t}*\delta)(\ve x) \nonumber \\
    &= \int \dynV{\ve x'}{t}\delta(\ve x - \ve x')d\nu(\ve{x}')  \label{eq:apply_dirac}
\end{align}
where $\delta(\cdot)$ is the Dirac function
$\delta(\ve{r}) = 0,\ \ve{r} \neq 0 $ and 
$\int \delta(\ve{r}) d\nu = 1$ and $\nu(\cdot)$ is the volume integration variable.

Replacing the $\delta(\cdot)$ by a kernel with finite support, $\delta \to W$, where $W$ is a kernel locally supported close to $0$ ($W(\ve x) = 0$ when $\|\ve{x}\|$ exceeds a small radius and $\int W(\ve{x})d\nu(\ve{x})=1$), the finite-width kernel approximation leads to a quantization scheme of a physical field using the notion of {\em particles} \cite{gingold1977smoothed}.
A particle represents the interesting physical quantities within a small spatial extent. In the NeRF problem, it is the {\em visual features} that are of the central interest, i.e., the vector quantity produced by the module V in \eqref{eq:feature_V_xt}. The visual feature is fed into subsequent MLPs to output the light radiance properties (density and color). In our model, the visual feature field is represented via smoothed particles \cite{gingold1977smoothed}. When querying a location $\ve{x}$ at time $t$,
\begin{align}
\dynVField{t}(\ve x) \approx \sum_{i = 1}^{N_p} \ve{v}^i W(\ve x- \ve {p}^i(t)) \label{eq:discretization}
\end{align}
where $i$ specifies one of the $N_p$ particles. As shown in Fig.~\ref{fig:hybrid_rep} (b), each particle represents the movements of a small finite volume of certain visual characteristics, consisting of $\ve p^i(t)$ and $\ve v^i$. $\ve p^i(t)$ is a 3D trajectory, mapping time $t$ to a 3D position. The latter, $\ve v^i$, is a time-invariant appearance feature.

Due to how the particles contribute to the NeRF model, we call them {\em appearance particles}. While our method and ParticleNeRF \cite{Chakra22_ParticleNeRF} share a foundational concept from [10], our approach employs significantly different computational details, which are presented in the next subsection..


\subsection{Appearance Particle Model\label{sec:ap}}
\begin{figure}
    \centering
    \includegraphics[width=0.75\columnwidth]{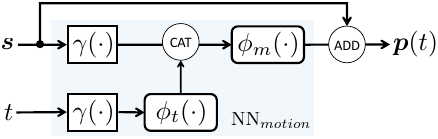}
    \caption{\label{fig:dynamics_model} Computational structure for time-varying position of a particle. $\gamma$ is the positional encoding function \cite{Mildenhall20}, $\phi_m$ and $\phi_t$ are $3$ and $2$-layer MLP, `CAT' is concatenate operation and `ADD' is element-wise addition.}
\end{figure}
\begin{figure}
    \centering
    \includegraphics[width=0.95\columnwidth]{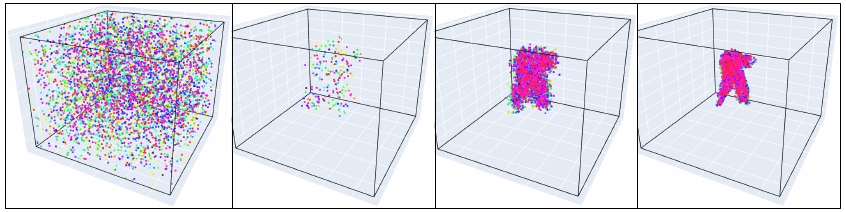}
    \caption{\label{fig:particles_in_training} Training process of particles at $t=0$. Panes show the particles at $t=0$ during different stages of training. (See Sec.~\ref{sec:ap}) }
\end{figure}

Recall that an individual particle carries two pieces of dynamics information, a visual feature $\ve {v}$ and time-varying position $\ve {p}(t)$. 
In DAP-NeRF, $\ve v^i$ is a per-particle learnable feature vector of appearance characteristics. $\ve p ^i(\cdot)$ is implemented using a small neural network, 
\begin{align}
    \ve p^i(t) := \textrm{NN}_{motion}(t, \ve{s}^i) + \ve{s}^i
    \label{eq:particle_motion_model}
\end{align}
where $\textrm{NN}_{motion}$ is the neural network that computes the particle motion as the offset from the starting point $\ve s^i$.
The computation is illustrated in Fig. \ref{fig:dynamics_model}. Therefore, the learning model of particles consists of i) per-particle visual feature $\ve {v}^i$, ii) per-particle starting point $\ve s^i$ and iii) the network parameters of $\textrm{NN}_{motion}$ that are shared by all particles.

The particles are designed to adaptively model the dynamic components using the following mechanism:

\paragraph{\textbf{Initialization}}
We initially place a set of particles by randomly distributing them within the pre-defined bounding box encompassing the scene. For each placed particle, we initialize $\ve s^i$ as the spatial coordinate. The setup is illustrated in the first pane of Fig. \ref{fig:overview}~(b).

\paragraph{\textbf{Particle Removal and Re-sampling}\label{sec:removal_resample}}
To ensure that particles represent only the dynamic components of a scene, we introduce a run-time strategy that automatically removes and re-samples particles during training. Specifically, we
\begin{itemize}
    \item remove particles that are located in the known free space or rarely move throughout the whole timeline. Free spaces are identified using the occupancy mask technique described in \cite{Sun22_DVGO}, i.e., alpha value less than a threshold $\epsilon_{\alpha}$. The immobility of a particle $i$ is identified by the length of its trajectory, i.e., $\int_0^1 \|\frac{\partial \ve{p}^i(r)}{\partial r}\|_2 dr$ not exceeding a threshold $\epsilon_\text{traj}$. The removal step is performed every few training steps from the beginning. The second pane of Fig. \ref{fig:particles_in_training} shows the result of the removal.
    \item conduct random re-sampling immediately after each removal step, aiming to preserve the overall particle count. New particles are randomly placed within a small radius around the remaining particles, and each initially inherits the visual feature $\ve {v}$ of the particle from which it was sampled. The third pane of Fig. \ref{fig:particles_in_training} displays the first re-sampling, and the fourth pane is the final result of alternate removal and re-sampling.
\end{itemize}

\subsection{Efficient Computation\label{sec:eff_compute}}
\paragraph{\textbf{Dynamic Feature Field}}
The computation of \eqref{eq:discretization} requires nearest neighbor search, which is expensive when the number of particles $N_p$ is large. We propose a computational scheme referring to the grid structure, inspired by the strategy of \cite{Sulsky1993APM}.

Specifically, we employ a grid $\grid{G}^{D}$ of $N_x \times N_y \times N_z$ nodes and each node is associated with a $C$-dimension feature. The node feature is of the same format as the particle visual feature $\ve v$.
Then the computation of $\dynV{\ve x}{t}$ consists of two steps: i) propagating appearance information from particles to grid nodes, and ii) passing the information to the query location. 

\newcommand{\weightin}[2]{w_{{#1}\to{#2}}}
\newcommand{\triweightin}[2]{w^{\mathrm{tri}}_{{#1}\to{#2}}}
At time $t$, we update features at $n$-th node of $\grid{G}^{D}$ by:
\begin{align}
\grid{G}^D(n) &= \sum_{i\in \mathcal N(n)} \weightin{i}{n} \ve{v}^i 
\label{eq:particle_to_grid}
\end{align}
where $\mathcal N(n)$ represents the particles in one of the cells adjacent to $n$.
The {\em distribution weight} $\weightin{i}{n}$ is the tri-linear interpolation weight for particle $i$ in the grid cell containing $n$. Having appearance features propagated to $\grid{G}^D$,  $\dynV{\ve x}{t}$ is computed via standard tri-linear interpolation
\begin{align}
\text{Interp}(\ve x, \grid{G}^{D}): (\mathbb{R}^3, \mathbb{R}^{C\times N_x \times N_y \times N_z}) \mapsto \mathbb{R}^C \label{eq:feat_field_trilinear}
\end{align}
We query this field by directly accessing neighbor grid nodes instead of searching neighbors over all particles like the previous method \cite{Chakra22_ParticleNeRF}. This approach reduces the computational complexity from O($N_p$) to O($1$). 

\paragraph{\textbf{Static Feature Field}}
For the static field component $V^{s}(\ve x)$, we adopt the canonical feature grid as in \cite{Fang22_TINeuVox}. The static feature grid $\grid{G}^S \in \mathbb{R}^{C\times N_x \times N_y \times N_z}$ shares the same configuration as $\grid{G}^D$, and is initialized with zero values throughout. The querying of this feature field is also done via tri-linear interpolation.

\paragraph{\textbf{Field Superposition}}
Since $\grid{G}^S$ and $\grid{G}^D$ are both grid-based and aligned in shape, the superposition of fields as described in \eqref{eq:superimpose} (for a convenient simplified reminder, consider $V:=(1-\alpha)V^s+\alpha V^d$) is straightforwardly implemented as
\begin{equation}
\begin{aligned}
\grid{G}(n) &= (1 - m(n)) \grid{G}^S(n) + m(n) \grid{G}^D(n) \\
m(n) &= 
    \begin{cases}
    1 & \text{if} \quad \sum_i^{N_p} \weightin{i}{n} >0 \\
    0 & \text{otherwise}
    \end{cases} \label{eq:superposition_weight}
\end{aligned}
\end{equation}
where $n$ denotes the node number, and $\weightin{i}{n}$ is the tri-linear interpolation weight for particle $i$ in the grid cell containing $n$. \eqref{eq:superposition_weight} means that the static field is active only when there is no particle in the neighboring grid cells.

We implement superpositional feature field via
\begin{equation}
\begin{aligned}
\ve{f}_v &= \text{Interp}(\ve{x}, \grid{G}) \\
V(\ve x) &= \phi_v(\gamma(\ve{f}_{v}), \gamma(\ve x))
\end{aligned}
\end{equation}
where $\phi_v$ is a small neural network detailed in Fig.~\ref{fig:nn_arch} (a), `Interp' denotes tri-linear interpolation.

\subsection{Optimization \label{sec:optim}}
\paragraph{\textbf{Pipeline}}
The training process of our model follows established practices in dynamic NeRFs \cite{Pumarola21_D-NeRF, Fang22_TINeuVox}, involving multiple iterations of model optimization. Each iteration begins by selecting an image from a video sequence. Spatial points are then sampled along rays that extend from the camera center to a batch of random pixel locations. Using the image’s timestamp, particle states are updated according to our methods detailed in Sec. \ref{sec:ap}. Subsequently, we create the superpositional radiance field, from which appearance information is queried using the sampled spatial points.

Finally, we employ the standard volume rendering \cite{Mildenhall20} on each ray to compute the colors of the pixels and render an image. To optimize the model, losses between the rendered and actual images are computed, as detailed in subsequent paragraphs.

\paragraph{\textbf{Losses}}
Following TiNeuVox \cite{Fang22_TINeuVox}, our training process involves selecting camera rays randomly, querying the radiance field, and applying standard volumetric rendering \cite{Mildenhall20} to compute ray colors and losses. Firstly, we adopt the standard photometric loss, quantified as the squared error between the rendered color $\mathbf C(\mathbf r)$ and the ground-truth color $\hat{\mathbf C}(\mathbf r)$:
\begin{align}
\mathcal L_\text{photo}=\Vert \hat{\mathbf C}(\mathbf r) - \mathbf C(\mathbf r) \Vert_2^2
\end{align}

Following \cite{Sun22_DVGO, Fang22_TINeuVox}, we apply two additional losses. The per-point RGB loss $\mathcal L_\text{ptrgb}$ stabilize the optimization process by supervising all sampled points with the target color. The background entropy loss $\mathcal L_\text{bg}$ encourages the model to better distinguish between foreground and background areas.

Since explicitly modeling dynamic elements is non-trivial, we further employ two regularization terms for the superpositional feature field and the motion model of particles. Specifically, we compute the total variation loss on the superpositional feature grid $\grid{G}$ using
\begin{equation}
\begin{aligned}
\mathcal L_\text{tvf} = \text{TV}(\grid{G}) := \frac{1}{N}\sum_{i,j,k} \Bigg( \left\| \grid{G}_{i,j,k} - \grid{G}_{i+1,j,k} \right\|_2 + \\
\left\| \grid{G}_{i,j,k} - \grid{G}_{i,j+1,k} \right\|_2 + \left\| \grid{G}_{i,j,k} - \grid{G}_{i,j,k+1} \right\|_2 \Bigg)
\end{aligned}
\end{equation}
where $N$ is the total number of grid nodes, and $\text{TV}(\cdot)$ is the operator calculating the total variation loss on a grid. Inspired by the As Rigid As Possible (ARAP) loss, the second regularization term aims to preserve local rigidity within the particles. Due to the costly nearest neighbor search required by the original ARAP loss, we employ an efficient computation method as introduced in Sec.~\ref{sec:eff_compute}. Specifically, we construct a \textit{motion grid} $\grid{G}^m$ where each node summarizes the movement of nearby particles via 
\begin{align}
\grid{G}^m(n) &= \frac{1}{w_n} \sum_{i\in \mathcal N(n)} \weightin{i}{n} \textrm{NN}_{motion}(t, \ve{s}^i)
\end{align}
where $w_n=\sum_{i\in \mathcal N(n)} \weightin{i}{n}$ is used for normalization. The term $\textrm{NN}_{motion}(t, \ve{s}^i)$ denotes particle offset at time $t$ as defined in \eqref{eq:particle_motion_model}. $\mathcal N(n)$ and $\weightin{i}{n}$ are defined as in \eqref{eq:particle_to_grid}. The motion regularization loss for particles is then implemented by
\begin{align}
\mathcal L_\text{tvm} = \text{TV}(\grid{G}^m)
\end{align}

Therefore, the overall loss for the model is defined as
\begin{align}
\mathcal L := \mathcal L_\text{photo} + w_1 \mathcal L_\text{ptrgb} + w_2 \mathcal L_\text{bg} + w_3 \mathcal L_\text{tvf} + w_4 \mathcal L_\text{tvm}  \label{eq:loss_items}
\end{align}


\paragraph{\textbf{Grid-based Coarse to Fine}}
Following the common practice in \cite{Sun22_DVGO, Fang22_TINeuVox}, all the grids ($\grid{G}^D$, $\grid{G}^S$ and $\grid{G}$) will increase their total voxel number from an initial value (e.g. $80^3$) to a final value (e.g. $160^3$) during training. Specifically, the bounding box (BBox) that tightly encloses the camera frustums of the training views is initially identified. The lengths of this BBox are denoted as $L_x$, $L_y$, and $L_z$, while the expected number of voxels is represented as $M$. The size of the grid is then established as $(L_x/s) \times (L_y/s) \times (L_z/s)$, where $s=\sqrt[3]{L_x \cdot L_y \cdot L_z / M}$.

\begin{table*}[htbp]
    \centering
    \caption{Per-scene quantitative comparisons on D-NeRF dataset \cite{Pumarola21_D-NeRF}.}
    \resizebox{\linewidth}{!}{
    \begin{tabular}{lcccccccccccc}
    \toprule
    \multicolumn{1}{c}{\multirow{2}{1.2cm}{\textbf{Method}}} & \multicolumn{3}{c}{Hell Warrior} & \multicolumn{3}{c}{Mutant} & \multicolumn{3}{c}{Hook} & \multicolumn{3}{c}{Bouncing Balls} \\
                & PSNR$\uparrow$ & SSIM$\uparrow$ & LPIPS$\downarrow$ & PSNR$\uparrow$ & SSIM$\uparrow$ & LPIPS$\downarrow$ & PSNR$\uparrow$ & SSIM$\uparrow$ & LPIPS$\downarrow$ & PSNR$\uparrow$ & SSIM$\uparrow$ & LPIPS$\downarrow$ \\
    \cmidrule(lr){1-1} \cmidrule(lr){2-4} \cmidrule(lr){5-7} \cmidrule(lr){8-10} \cmidrule(lr){11-13}
    
    T-NeRF \cite{Pumarola21_D-NeRF}     & 23.19     & 0.93      & 0.08      & 30.56     & 0.96      & 0.04      & 27.21     & 0.94      & 0.06      & 37.81     & 0.98      & 0.12      \\
    D-NeRF \cite{Pumarola21_D-NeRF}     & 25.02     & 0.95      & 0.06      & 31.29     & 0.97      & 0.02      & 29.25     & 0.96      & 0.06      & 38.93     & 0.98      & 0.10      \\
    NDVG \cite{Guo22_NDVG}              & 25.53     & 0.95      & 0.07     & 35.53     & \textbf{0.99}  & \textbf{0.01}     & 29.80     & 0.97      & \textbf{0.04}     & 34.58     & 0.97      & 0.11     \\
    TiNeuVox \cite{Fang22_TINeuVox}     & 28.17     & \textbf{0.97}  & 0.07      & 33.61     & 0.98      & 0.03      & 31.45     & 0.97      & 0.05      & 40.73     & \textbf{0.99}  & \textbf{0.04}      \\
    Tensor4D \cite{Tensor4D} & - & - & - & - & - & - & - & - & - & - & - & - \\
    K-Planes \cite{Keil_K-Planes}       & 24.81     & 0.95      &  -        & 32.59     & 0.97      &  -        & 28.13     & 0.95      &  -        & 40.33     & \textbf{0.99}  &  -        \\
    DAP-NeRF (ours)       & \textbf{29.51} & \textbf{0.97}  &  \textbf{0.05}        & \textbf{35.75} & \textbf{0.99}  &  \textbf{0.01}        & \textbf{32.69} & \textbf{0.98}  &  \textbf{0.04}      & \textbf{41.29} & \textbf{0.99}  &  \textbf{0.04}        \\

    \midrule

    \multicolumn{1}{c}{\multirow{2}{1.2cm}{\textbf{Method}}} & \multicolumn{3}{c}{Lego} & \multicolumn{3}{c}{T-Rex} & \multicolumn{3}{c}{Stand Up} & \multicolumn{3}{c}{Jumping Jacks} \\
                & PSNR$\uparrow$ & SSIM$\uparrow$ & LPIPS$\downarrow$ & PSNR$\uparrow$ & SSIM$\uparrow$ & LPIPS$\downarrow$ & PSNR$\uparrow$ & SSIM$\uparrow$ & LPIPS$\downarrow$ & PSNR$\uparrow$ & SSIM$\uparrow$ & LPIPS$\downarrow$ \\
    \cmidrule(lr){1-1} \cmidrule(lr){2-4} \cmidrule(lr){5-7} \cmidrule(lr){8-10} \cmidrule(lr){11-13}

    T-NeRF \cite{Pumarola21_D-NeRF}     & 23.82     & 0.90      & 0.15      & 30.19     & 0.96      & 0.13      & 31.24     & 0.97      & 0.02      & 32.01     & 0.97      & 0.03      \\
    D-NeRF \cite{Pumarola21_D-NeRF}     & 21.64     & 0.83      & 0.16      & 31.75     & 0.97      & 0.03      & 32.79     & 0.98      & 0.02      & 32.80     & 0.98      & 0.03      \\
    NDVG \cite{Guo22_NDVG}              & 25.23     & 0.93      & \textbf{0.05}     & 30.15     & 0.97      & 0.05     & 34.05     & 0.98      & 0.02     & 29.45     & 0.96      & 0.08     \\
    TiNeuVox \cite{Fang22_TINeuVox}     & 25.02     & 0.92      & 0.07      & 32.70     & \textbf{0.98}  & 0.03      & 35.43     & \textbf{0.99}  & 0.02      & 34.23     & 0.98      & 0.03      \\
    Tensor4D \cite{Tensor4D} & \textbf{26.71} & \textbf{0.95} & \textbf{0.03} & - & - & - & 36.32 & 0.98 & 0.02 & 34.43 & 0.98 & 0.03 \\
    K-Planes \cite{Keil_K-Planes}       & 25.27     & 0.94      &  -        & 30.75     & 0.97      &  -        & 33.17     & 0.98      &  -        & 31.64     & 0.97      &  -        \\
    DAP-NeRF (ours)       & 25.43 & 0.94  &  0.05        & \textbf{34.07} & \textbf{0.98}  &  \textbf{0.02}        & \textbf{37.86} & \textbf{0.99}  &  \textbf{0.01}        & \textbf{35.90} & \textbf{0.99}  &  \textbf{0.02}        \\
    
    \bottomrule
    \end{tabular}
    }
    \label{tab:per_scene_NVS}
\end{table*}

\begin{table*}[htbp]
    \centering
    \caption{Per-scene quantitative comparisons on NHR dataset \cite{nhr}.}
    \resizebox{\linewidth}{!}{
    \begin{tabular}{lcccccccccccc}
    \toprule
    \multicolumn{1}{c}{\multirow{2}{1.2cm}{\textbf{Method}}} & \multicolumn{3}{c}{Sport 1} & \multicolumn{3}{c}{Sport 2} & \multicolumn{3}{c}{Sport 3} & \multicolumn{3}{c}{Basketball}  \\
                & PSNR$\uparrow$ & SSIM$\uparrow$ & LPIPS$\downarrow$ & PSNR$\uparrow$ & SSIM$\uparrow$ & LPIPS$\downarrow$ & PSNR$\uparrow$ & SSIM$\uparrow$ & LPIPS$\downarrow$ & PSNR$\uparrow$ & SSIM$\uparrow$ & LPIPS$\downarrow$
    \\
    \cmidrule(lr){1-1} \cmidrule(lr){2-4} \cmidrule(lr){5-7} \cmidrule(lr){8-10} \cmidrule(lr){11-13}
    
    TiNeuVox-S \cite{Fang22_TINeuVox} & 27.82 & 0.972 & 0.089 & 27.84 & 0.975 & 0.062 & 27.45 & 0.968 & 0.089 & 25.46 & 0.953 & 0.011 \\
    TiNeuVox-B \cite{Fang22_TINeuVox} & 28.07 & 0.974 & 0.082 & 28.22 & 0.978 & 0.070 & 28.29 & 0.974 & 0.084 & 26.31 & 0.962 & 0.010 \\
    
    DAP-NeRF (ours)                   & \textbf{28.58} & \textbf{0.978} & \textbf{0.071} & \textbf{28.78} & \textbf{0.981} & \textbf{0.067} & \textbf{28.50} & \textbf{0.976} & \textbf{0.078} & \textbf{26.41} & \textbf{0.964} & \textbf{0.090} \\
    
    \bottomrule
    \end{tabular}
    }
    \label{tab:per_scene_NVS_nhr}
\end{table*}

\definecolor{darkred}{rgb}{0.6, 0.0, 0.0}
\definecolor{darkblue}{rgb}{0.0, 0.0, 0.6}
\newcommand{\bestresult}[1]{{\color{darkred}\textbf{{#1}}}}
\newcommand{\secondbest}[1]{{\color{darkblue}\textbf{{#1}}}}

\begin{table*}[htbp]
    \centering
    \caption{Per-scene quantitative comparisons on HyperNeRF dataset \cite{Park21_HyperNeRF}. \bestresult{Red} text indicates the best result while \secondbest{blue} text indicate the second best.}
    \resizebox{\linewidth}{!}{
    \begin{tabular}{lcccccccccccc}
    \toprule
    \multicolumn{1}{c}{\multirow{2}{1.2cm}{\textbf{Method}}} & \multicolumn{3}{c}{3D Printer} & \multicolumn{3}{c}{Broom} & \multicolumn{3}{c}{Chicken} & \multicolumn{3}{c}{Peel Banana}  \\
        & PSNR$\uparrow$ & SSIM$\uparrow$ & LPIPS$\downarrow$ & PSNR$\uparrow$ & SSIM$\uparrow$ & LPIPS$\downarrow$ & PSNR$\uparrow$ & SSIM$\uparrow$ & LPIPS$\downarrow$ & PSNR$\uparrow$ & SSIM$\uparrow$ & LPIPS$\downarrow$
    \\
    \cmidrule(lr){1-1} \cmidrule(lr){2-4} \cmidrule(lr){5-7} \cmidrule(lr){8-10} \cmidrule(lr){11-13}
    
    NeRF \cite{Mildenhall20} & 20.7 & 0.780 & - & 19.9 & 0.653 & - & 19.9 & 0.777 & - & 20.0 & 0.769 & - \\
    Nerfies \cite{ParkSBBGSM21_Nerfies} & 20.6 & 0.830 & - & 19.2 & 0.567 & - & 26.7 & 0.943 & - & 22.4 & 0.872 & -  \\
    HyperNeRF \cite{Park21_HyperNeRF} & 20.0 & 0.821 & - & 19.3 & 0.591 & - & 26.9 & \bestresult{0.948} & - & \secondbest{23.3} & \bestresult{0.896} & - \\
    NDVG \cite{Guo22_NDVG} & 22.4 & 0.839 & - & \secondbest{21.5} & \secondbest{0.703} & - & 27.1 & 0.939 & - & 22.8 & 0.828 & - \\
    TiNeuVox-B \cite{Fang22_TINeuVox} & \secondbest{22.8} & \secondbest{0.841} & - & \secondbest{21.5} & 0.686 & - & \bestresult{28.3} & \secondbest{0.947} & - & \bestresult{24.4} & \secondbest{0.873} & - \\
    
    DAP-NeRF (ours) & \bestresult{22.9} & \bestresult{0.845} & 0.161 & \bestresult{21.9} & \bestresult{0.713} & 0.175 & \secondbest{27.6} & 0.939 & 0.066 & 22.5 & 0.811 & 0.137 \\
    
    \bottomrule
    \end{tabular}
    }
    \label{tab:per_scene_NVS_hyper}
\end{table*}

\begin{table*}[htbp]
    \centering
    \caption{Per-scene quantitative comparisons on NeRF-DS dataset \cite{Yan23_NeRF-DS}. \bestresult{Red} text indicates the best result while \secondbest{blue} text indicate the second best.}
    \resizebox{\linewidth}{!}{
    \begin{tabular}{lcccccccccccc}
    \toprule
    \multicolumn{1}{c}{\multirow{2}{1.2cm}{\textbf{Method}}} & \multicolumn{3}{c}{Sieve} & \multicolumn{3}{c}{Plate} & \multicolumn{3}{c}{Bell} & \multicolumn{3}{c}{Press} \\
                & PSNR$\uparrow$ & SSIM$\uparrow$ & LPIPS$\downarrow$ & PSNR$\uparrow$ & SSIM$\uparrow$ & LPIPS$\downarrow$ & PSNR$\uparrow$ & SSIM$\uparrow$ & LPIPS$\downarrow$ & PSNR$\uparrow$ & SSIM$\uparrow$ & LPIPS$\downarrow$ \\
    \cmidrule(lr){1-1} \cmidrule(lr){2-4} \cmidrule(lr){5-7} \cmidrule(lr){8-10} \cmidrule(lr){11-13}
    
    TiNeuVox \cite{Fang22_TINeuVox}         & 21.49     & 0.8265    & 0.3176    & \secondbest{20.58}     & 0.8027    &  0.3317   & 23.08     & \secondbest{0.8242}    & 0.2568    & 24.47     & 0.8613   & 0.3001    \\
    HyperNeRF \cite{Park21_HyperNeRF}       & 25.43     & \secondbest{0.8798}    & 0.1645    & 18.93     & 0.7709    &  0.2940   & 23.06     & 0.8097    & 0.2052    & \secondbest{26.15}     & \secondbest{0.8897}   & \secondbest{0.1959}    \\
    NeRF-DS \cite{Yan23_NeRF-DS}            & \bestresult{25.78}     & \bestresult{0.8900}    & \bestresult{0.1472}    & 20.54     & \secondbest{0.8042}    &  \bestresult{0.1996}   & \secondbest{23.19}     & 0.8212    & \secondbest{0.1867}    & 25.72     & 0.8618   & 0.2047    \\
    DAP-NeRF (ours)                         & \secondbest{25.61}     & 0.8698    & \secondbest{0.1603}    & \bestresult{20.66}     & \bestresult{0.8135}    &  \secondbest{0.2374}   & \bestresult{24.11}     & \bestresult{0.8334}    & \bestresult{0.1641}    & \bestresult{26.18}     & \bestresult{0.8932}   & \bestresult{0.1920}    \\

    \midrule

    \multicolumn{1}{c}{\multirow{2}{1.2cm}{\textbf{Method}}} & \multicolumn{3}{c}{Cup} & \multicolumn{3}{c}{As} & \multicolumn{3}{c}{Basin} & \multicolumn{3}{c}{Mean} \\
                & PSNR$\uparrow$ & SSIM$\uparrow$ & LPIPS$\downarrow$ & PSNR$\uparrow$ & SSIM$\uparrow$ & LPIPS$\downarrow$ & PSNR$\uparrow$ & SSIM$\uparrow$ & LPIPS$\downarrow$ & PSNR$\uparrow$ & SSIM$\uparrow$ & LPIPS$\downarrow$ \\
    \cmidrule(lr){1-1} \cmidrule(lr){2-4} \cmidrule(lr){5-7} \cmidrule(lr){8-10} \cmidrule(lr){11-13}

    TiNeuVox \cite{Fang22_TINeuVox}         & 19.71     & 0.8109    & 0.3643    & 21.26     & 0.8289    &  0.3967   & \secondbest{20.66}     & 0.8145    & 0.2690    & 21.61     & 0.8234   & 0.2766    \\
    HyperNeRF \cite{Park21_HyperNeRF}       & 24.59     & \secondbest{0.8770}    & \secondbest{0.1650}    & \secondbest{25.58}     & \bestresult{0.8949}    &  \secondbest{0.1777}   & 20.41     & \secondbest{0.8199}    & \secondbest{0.1911}    & 23.45     & 0.8488   & 0.1990    \\
    NeRF-DS \cite{Yan23_NeRF-DS}            & \bestresult{24.91}     & 0.8741    & 0.1737    & 25.13     & 0.8778    &  \bestresult{0.1741}   & 19.96     & 0.8166    & \bestresult{0.1855}    & \secondbest{23.60}     & \secondbest{0.8494}   & \bestresult{0.1816}    \\
    DAP-NeRF (ours)                         & \secondbest{24.82}     & \bestresult{0.8876}    & \bestresult{0.1608}    & \bestresult{25.83}     & \secondbest{0.8802}    &  0.1814   & \bestresult{20.72}     & \bestresult{0.8211}    & 0.2134    & \bestresult{23.99}     & \bestresult{0.8570}   & \secondbest{0.1871}    \\
    
    \bottomrule
    \end{tabular}
    }
    \label{tab:per_scene_NVS_nerfds}
\end{table*}


\begin{figure}[tbp]
    \centering
    \includegraphics[width=0.7\columnwidth]{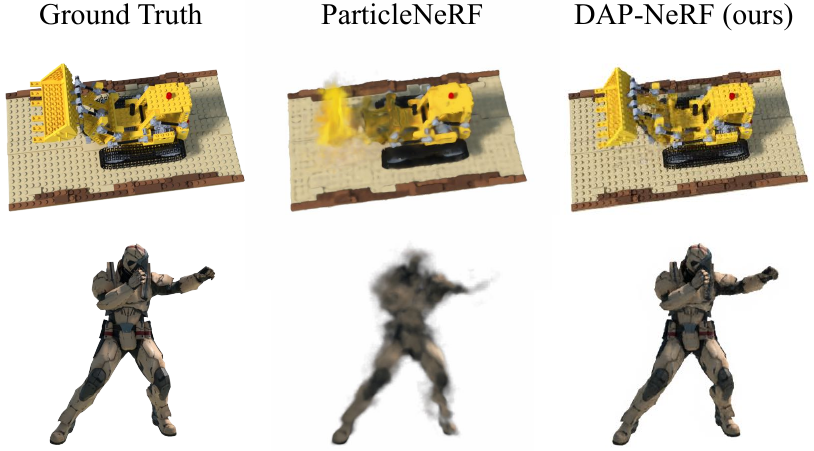}
    \caption{\label{fig:particlenerf_test} 
    Comparative analysis with ParticleNeRF \cite{Chakra22_ParticleNeRF} on the D-NeRF dataset \cite{Pumarola21_D-NeRF}. Although both methods utilize particle-based representations, ParticleNeRF struggles with monocular video data. 
    }
\end{figure}

\begin{figure}[tbp]
    \centering
    \includegraphics[width=0.8\columnwidth]{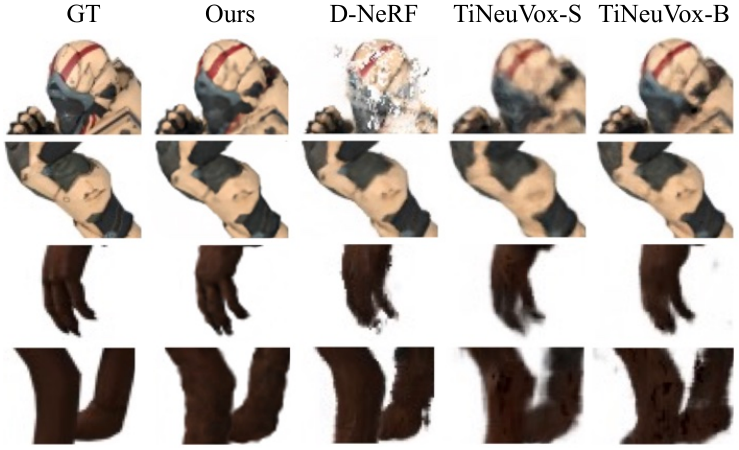}
    \caption{\label{fig:quality_compare_dnerf} Qualitative comparisons on D-NeRF dataset \cite{Pumarola21_D-NeRF}.
    }
\end{figure}

\begin{figure*}[htbp]
    \centering
    \includegraphics[width=0.99\textwidth]{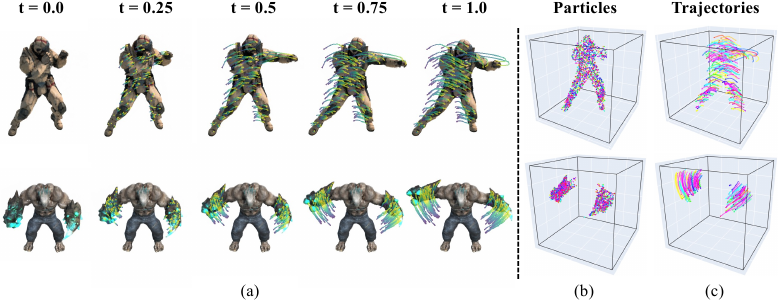}
    \caption{\label{fig:render_res} Synthesized novel views for D-NeRF dataset \cite{Pumarola21_D-NeRF} and particle-based dynamic models.
    {\bf(a)} displays the images rendered by the learned DAP-NeRF models, along with the trajectories of $100$ sampled particles drawn onto the images.
    {\bf (b)} shows the particles at time $t=0$. To enhance visual clarity, we only render $2$k randomly sampled particles.
    {\bf (c)} shows the corresponding learned trajectories of particles, where only $200$ randomly sampled particles are rendered.
    }
\end{figure*}
\begin{figure*}[htbp]
    \centering
    \includegraphics[width=0.99\textwidth]{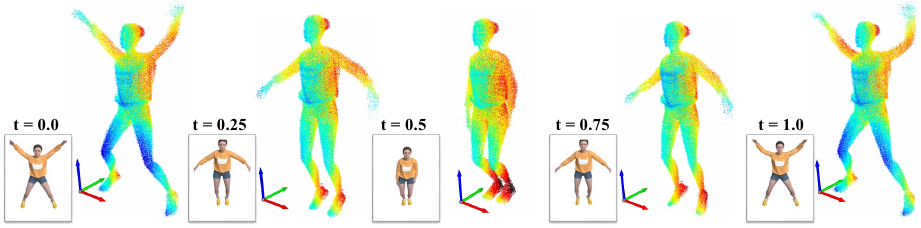}
    \caption{\label{fig:particle_vis_jump} Particle positions learned by DAP-NeRF at different frames. The color of each point is determined by its y-axis value.
    }
\end{figure*}

\begin{figure}[htbp]
    \centering
    \includegraphics[width=0.85\columnwidth]{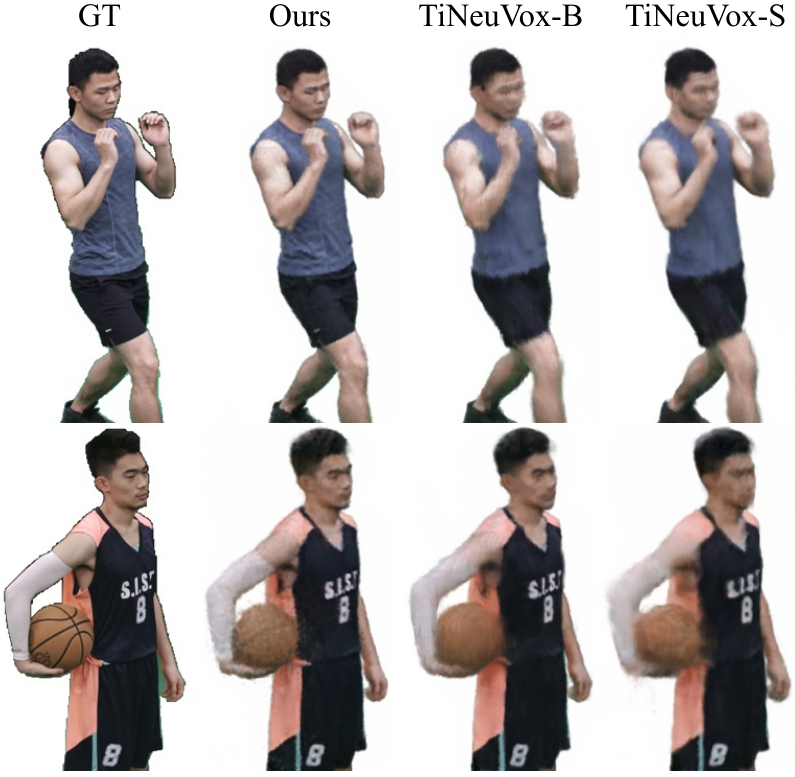}
    \caption{\label{fig:quality_compare_nhr} Qualitative comparisons on NHR dataset \cite{nhr}.
    }
\end{figure}

\begin{figure}[htbp]
    \centering
    \includegraphics[width=0.95\columnwidth]{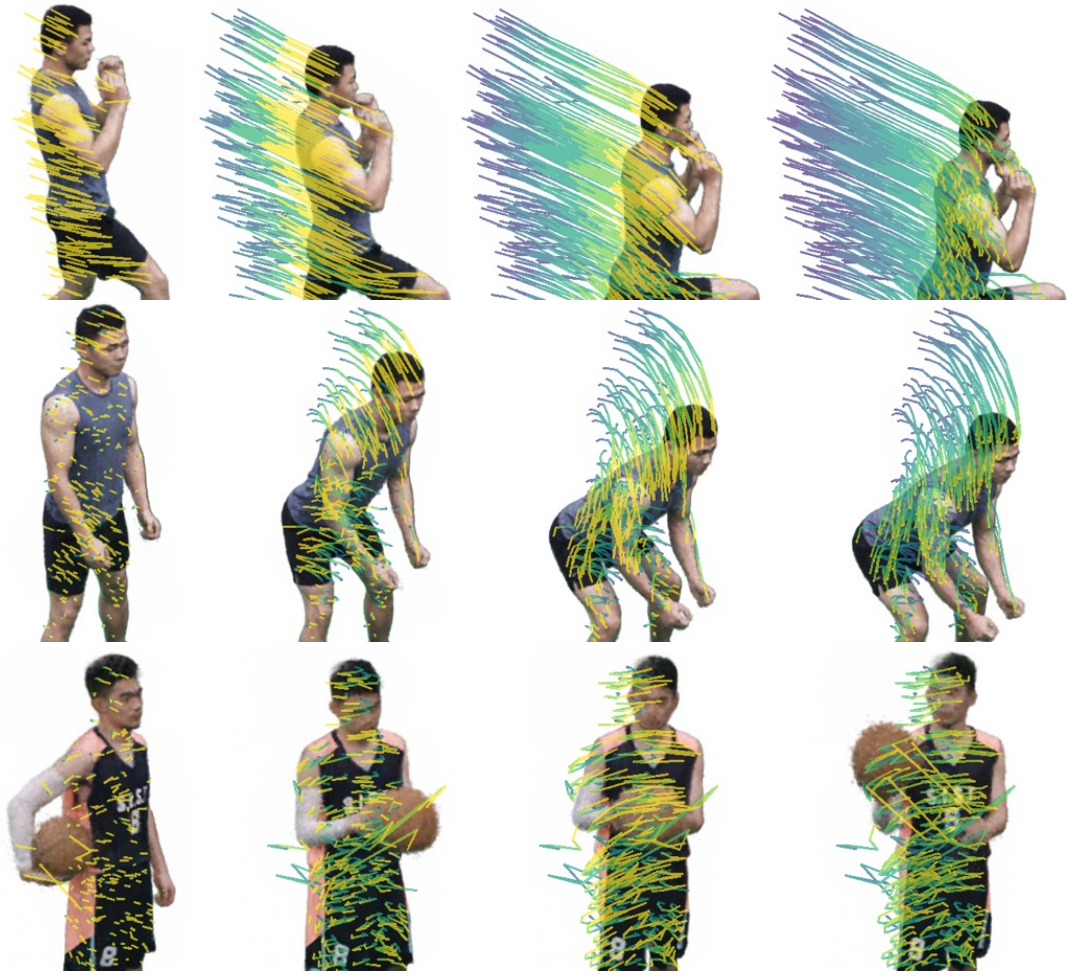}
    \caption{\label{fig:render_N_traj_nhr} This figure displays the novel views rendered by the learned DAP-NeRF models, along with the trajectories of $200$ randomly sampled particles.
    }
\end{figure}

\begin{figure*}[htbp]
    \centering
    \includegraphics[width=0.75\textwidth]{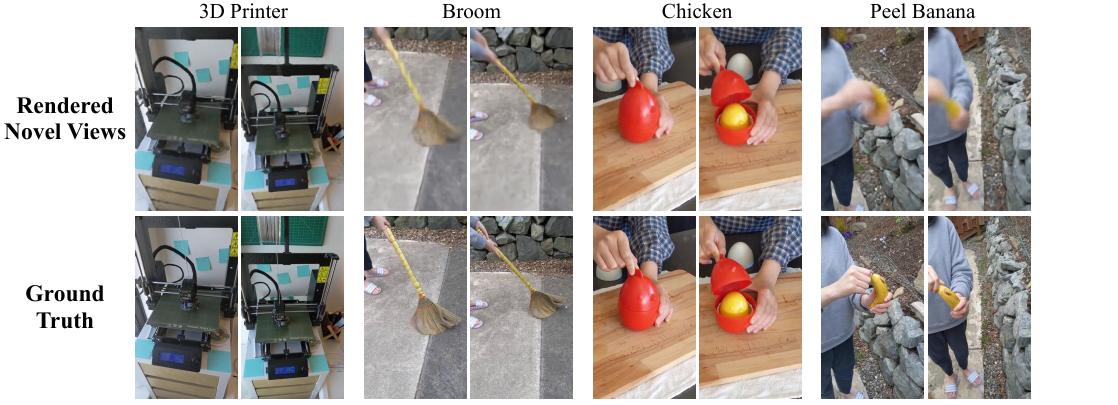}
    \caption{\label{fig:qualitative_result_hyper} Novel views rendered by DAP-NeRF on the HyperNeRF dataset \cite{Park21_HyperNeRF}. For each scene, we render images from two perspectives at various times.
    }
\end{figure*}

\section{Experiments}
After testing the validity of the proposed DAP-NeRF in novel view synthesis tasks, we demonstrate the capacity and advantage of the particle-based representation in motion modeling, as well as related applications. This section also explores ablation studies on particle quantity effects.

\subsection{Experimental Settings}
\begin{figure*}
    \centering
    \includegraphics[width=0.68\textwidth]{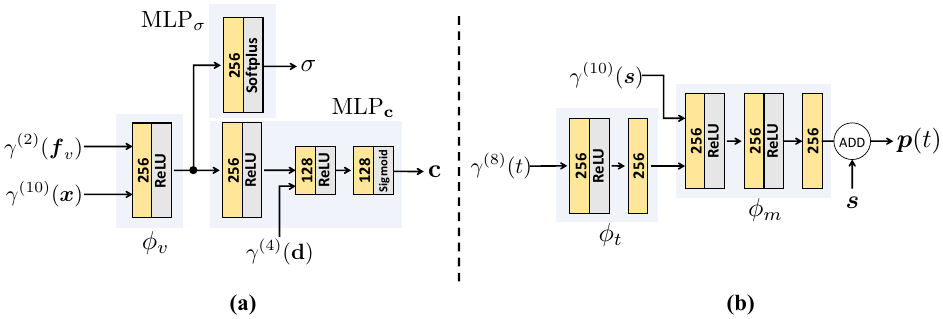}
    \caption{\label{fig:nn_arch} Network Architectures. {\bf (a)} shows the implementation of Fig. \ref{fig:common_dyn_nerf}, which is the NeRF architecture used in our method. {\bf (b)} details Fig. \ref{fig:dynamics_model}, which is the motion model for time-varying position of a particle. In this figure, $\phi_t$, $\phi_m$, $\phi_v$, $\textrm{MLP}_{\sigma}$ and $\textrm{MLP}_{\mathbf{c}}$ are neural networks. The superscript of $\gamma^{(\cdot)}$ is the frequency number of positional encoding. Yellow boxes denote linear layers, grey boxes denote activation functions.}
\end{figure*}

\paragraph{\textbf{Datasets}}
We evaluate our method using four datasets: one synthetic dataset, two real-scene datasets, and our specifically constructed testing scenarios.  The \textit{D-NeRF Dataset} \cite{Pumarola21_D-NeRF} includes eight 360° synthetic Scenes. For a fair comparison, we follow the settings of \cite{Fang22_TINeuVox} where images are trained and rendered at a resolution of $400 \times 400$. For real-scene evaluation, the \textit{NHR Dataset} \cite{nhr} offers four scenes with a single human performing a specific action in each, captured using a multi-camera dome system. We conduct evaluations on $100$ frames from each scene, allocating $90\%$ of the camera views for training and the remaining $10\%$ for testing. Another dataset for real scenes is provided by \textit{HyperNeRF} \cite{Park21_HyperNeRF}, which employs two phones to capture four real unbounded scenes. We also use \textit{NeRF-DS dataset} \cite{Yan23_NeRF-DS}, where the scenes contain moving and deforming specular objects.

Additionally, to comprehensively evaluate models in motion modeling, we've designed three test scenes and built a dataset of 300 images per scene. The dataset also includes data for ground-truth material motion at specific locations and times. We will provide further details in Sec. \ref{sec:motion_measure}.

\paragraph{\textbf{Model and Optimization Configuration}}
We implement our framework using PyTorch \cite{Paszke19_pytorch}. Following \cite{Fang22_TINeuVox}, the resolution of static feature grid $\grid{G}^S$ and dynamic feature grid $\grid{G}^D$ are set to $160^3$. We utilize $200$k particles for synthetic datasets, and $500$k for real-scene datasets. The dimensions $C$ of both particle and voxel features are set to $12$. Details of the network architecture are shown in Fig.~\ref{fig:nn_arch}, where the network width is $256$, the frequency number of positional encoding is set to 10 for $(x,y,z)$, $4$ for view direction $\ve{d}$, and $8$ for time $t$. Particle removal and re-sampling are performed every $2$k training steps. During removal, we set the thresholds for alpha value and trajectory length to $\epsilon_\alpha=0.0001$ and $\epsilon_\text{traj}=0.1$, respectively. The small radius of the range for random re-sampling is set to $0.1$ times the voxel length of the used feature grids.

For the optimization process, we employ the Adam optimizer \cite{Kingma14_adam}, configured with beta values of $(0.9, 0.99)$ and a minibatch size of $4,096$ rays. The learning rates are set to $0.005$ for particle features and $0.001$ for motion predictor network $\phi_m$. Other learning rates for static feature grid $\grid{G}^S$, radiance network $\denf{\cdot}$ and $\clrf{\cdot}$ are following \cite{Fang22_TINeuVox}. The total number of training iterations is $60$k. All the learning rates will finally decay by $0.1$ with an exponential schedule. The weights for the loss items in \eqref{eq:loss_items} are set to $w_1=0.01$, $w_2=0.001$, $w_3=0.01$, and $w_4=0.01$.

\subsection{Novel View Synthesis}
We begin by comparing ParticleNeRF \cite{Chakra22_ParticleNeRF} with our proposed DAP-NeRF on the monocular D-NeRF dataset, since both methods employ a particle-based representation but with fundamentally different implementations. Fig. \ref{fig:particlenerf_test} illustrates that our method successfully adapts to challenging monocular video data, while ParticleNeRF struggles with this setup. This highlights the superiority of our technical design in terms of both performance and application potential.

To further validate the effectiveness of our method, we conduct experiments on the novel view synthesis task and compare DAP-NeRF with state-of-the-art methods. Tab.~\ref{tab:per_scene_NVS} presents the quantitative results on the \textit{D-NeRF Dataset}, evaluated based on peak signal-to-noise ratio (PSNR), structural similarity (SSIM) \cite{Wang04_SSIM} and learned perceptual image patch similarity (LPIPS) \cite{Zhang18_LPIPS}. Fig.~\ref{fig:quality_compare_dnerf} provides qualitative comparisons, showcasing image details rendered by different methods.

Results demonstrate that DAP-NeRF matches or surpasses the performance of state-of-the-art models, even though our particle-based representation is primarily designed for explicit motion modeling rather than solely focusing on appearance accuracy. Fig. \ref{fig:render_res} (a) (b) and (c) illustrate the learned particles and their trajectories, which effectively capture and track the dynamic elements within the scenes. Furthermore, Fig.~\ref{fig:particle_vis_jump} offers a visual inspection of the scene geometry discovered by particles at different moments.

For the real scenes in \textit{NHR Dataset}, our method also achieves superior performance in both quantitative comparisons (see Tab.~\ref{tab:per_scene_NVS_nhr}) and qualitative comparisons (see Fig.~\ref{fig:quality_compare_nhr}). In Fig.~\ref{fig:render_N_traj_nhr}, we visualize the rendered results as well as the particle trajectories, which visually demonstrate high quality and also reveal the potential applications such as human motion capture. In the \textit{HyperNeRF Dataset}, our method achieves competitive performance (see Tab.~\ref{tab:per_scene_NVS_hyper} and Fig.~\ref{fig:qualitative_result_hyper}), although the improvements are not as significant as in the previous two datasets. These results can largely be attributed to the dataset's sparse, forward-facing camera setup, which limits the geometric constraints essential for explicit motion modeling. In the \textit{NeRF-DS Dataset} (see Tab.~\ref{tab:per_scene_NVS_nerfds}), our method achieves comparable or superior performance to the baseline NeRF-DS \cite{Yan23_NeRF-DS}, which was specifically designed for modeling dynamic specular objects.

Note that the core novelty of our method lies in the design of a particle-based representation, primarily aimed at enhancing motion modeling rather than solely improving rendering quality. In the subsequent sections, we will demonstrate our method's unique advantages in faster rendering (Sec.~\ref{sec:efficiency}), explicit motion modeling (Sec.~\ref{sec:motion_measure}), and its potential practical applications (Sec.~\ref{sec:applications}).

\subsection{User Study\label{sec:user_study}}

\begin{figure}[htbp]
    \centering
    \includegraphics[width=0.7\columnwidth]{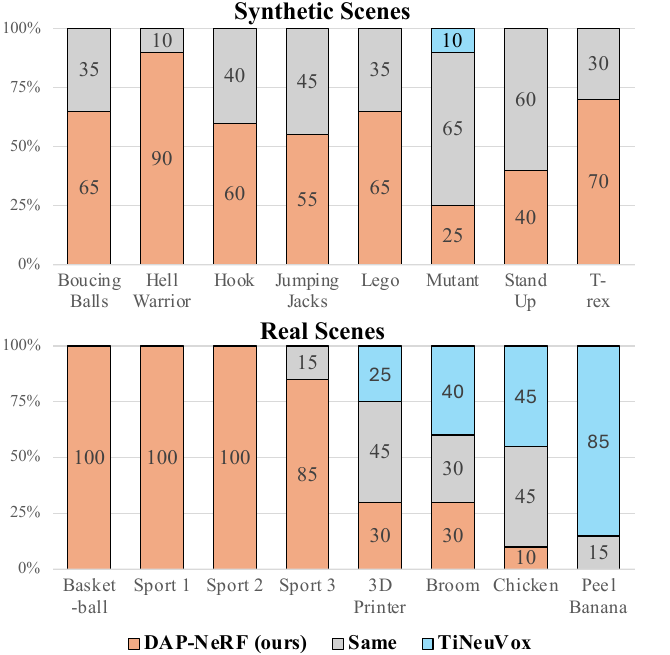}
    \caption{\label{fig:user_study}User study on rendering quality. This boxplot records the ratio of user preferences between DAP-NeRF (ours) and TiNeuVox \cite{Fang22_TINeuVox}. Participants could also select `Same' to indicate no significant difference between the methods.}
\end{figure}

To further evaluate reconstruction quality from a human perspective, we conducted a user study comparing our model's rendering performance to TiNeuVox \cite{Fang22_TINeuVox}, which is closest to ours in terms of PSNR and SSIM quantitative results. For each of the 16 scenes included in the D-NeRF, NHR, and HyperNeRF datasets, participants were presented with 4 random images rendered by both our method and TiNeuVox, alongside the corresponding ground-truth images. They were asked to judge which rendering was closer to the ground truth in terms of detail and overall visual accuracy, or if both renderings were of comparable quality. 

We collected questionnaires from 20 participants and compiled the results into a boxplot, as illustrated in Fig.~\ref{fig:user_study}. This user study yielded results consistent with the qualitative evaluations, further confirming the effectiveness of our method.

\subsection{Evaluation of Efficiency \label{sec:efficiency}}

\begin{table}[htbp]
    \centering
    \caption{Comparisons in terms of training and rendering times, and parameter counts on the D-NeRF Dataset \cite{Pumarola21_D-NeRF}.}
    \begin{tabular}{l|ccc}
    \toprule
    \textbf{Method} & \textbf{Train} & \textbf{Render (one frame)} & \textbf{\#.Params} \\
    \midrule
    D-NeRF \cite{Pumarola21_D-NeRF}   & 27.5 hours & 8.70s  & \textbf{1M} \\
    NDVG \cite{Guo22_NDVG}            & 23 mins    & 0.43s  & 87M \\
    TiNeuVox \cite{Fang22_TINeuVox}   & 28 mins    & 0.31s  & 25M \\
    K-Planes \cite{Keil_K-Planes}     & 49 mins    & 0.64s    & 37M \\
    DAP-NeRF (ours)                   & \textbf{17 mins} & \textbf{0.22s} & 52M \\
    \bottomrule
    \end{tabular}
    \label{tab:efficiency}
\end{table}

We compare our model with existing methods in terms of training and rendering times, as well as the total number of parameters. For a fair comparison, we evaluate all methods on the same device (RTX 3090 GPU) and report the average metrics across all scenes of the D-NeRF dataset \cite{Pumarola21_D-NeRF}.

As shown in Tab.~\ref{tab:efficiency}, our method achieves improved efficiency in both training and rendering times. Unlike methods that deform all ray-casting points (exceeding 30M points) to the canonical space, our method models the dynamics of only 0.2M particles, resulting in reduced computation. On the other hand, our model requires a reasonable increase in parameters to store particle features.

\newcommand{\deformf}{\textrm{df}}
\subsection{Evaluation of Motion modeling \label{sec:motion_measure}}
\begin{figure}[htbp]
    \centering
    \includegraphics[width=0.7\columnwidth]{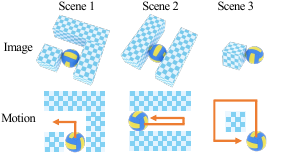}
    \caption{\label{fig:new_dataset} The dataset for evaluating motion modeling.}
\end{figure}

\begin{table}[htbp]
    \centering
    \caption{Motion Field Error (MFE) for evaluating motion modeling. The definition of MFE can be found in \eqref{eq:mfe}.}
    \begin{tabular}{l|ccc}
    \toprule
    \textbf{Method} & \textbf{Scene 1} & \textbf{Scene 2} & \textbf{Scene 3} \\
    \midrule
    TiNeuVox \cite{Fang22_TINeuVox} & 0.0197  & 0.0363 & 0.0376 \\
    DAP-NeRF (ours) & \textbf{0.0029} & \textbf{0.0104} & \textbf{0.0136} \\
    
    \bottomrule
    \end{tabular}
    
    \label{tab:motion_eval}
\end{table}

To further evaluate how well our models capture the physically relevant dynamics of a scene, we developed a dataset with explicit object motion information. As shown in Fig. \ref{fig:new_dataset}, each scene contains a moving object of interest (a ball), demonstrating varying levels of motion complexity.

Since the scene and object trajectories are constructed in a 3D design software \cite{Blender}, the ground-truth motions are known. Specifically, for any 3D coordinate $(x, y, z)$ at time $t$, we can determine if it's occupied and, if so, consider the occupying object's velocity as the velocity at $(x,y,z,t)$. With the ground-truth velocity, we design a metric to quantitatively evaluate the motion modeling. Formally, {\em Motion Field Error} (MFE) is defined as the average Euclidean-norm difference between velocities:
\begin{equation}
\begin{aligned}
    \textrm{MFE}(\vec F_1, \vec F_2) &:= \frac{1}{\mathcal V} \int_{\mathcal V} \Vert \vec F_1(\boldsymbol x) - \vec F_2(\boldsymbol x)) \Vert_2 dv \label{eq:mfe}
    \\
    & \approx \frac{1}{N} \sum_n^N \Vert \vec F_1(\boldsymbol x_n) - \vec F_2(\boldsymbol x_n)) \Vert_2
\end{aligned}
\end{equation}
where $\vec F_1, \vec F_2$ are the velocity fields to be compared, $\mathcal V$ is the volume of the region of interest. The integration \eqref{eq:mfe} is approximated by traversing the $N$ voxels, where $\ve{x}_n$ represents the center of a voxel. i) For existing deformable dynamic NeRF models, e.g. TiNeuVox \cite{Fang22_TINeuVox}, the velocity at a voxel is approximately implied the deformation field
\begin{align}
    \vec F_{\deformf}(\ve{x}) \approx \frac{\deformf(\ve{x}, t) - \deformf(\ve{x}, t + \delta t)}{\delta t}
\end{align}
where $\deformf$ stands for the deformation field component of a deformable dynamic NeRF. Given a time $t$, $\deformf$ specifies a 3D offset from a 3D location to a location in the canonical 3D model $\deformf: \mathbb R^3 \times \mathbb R^+ \mapsto \mathbb R^3$. Note that we need to zero-out the velocities in empty areas where deformable NeRFs are likely to produce incorrect values, as discussed in Sec. \ref{sec:ablation}. ii) In DAP-NeRF, the velocity field can be computed directly using the particle motions. Specifically, we computing the average velocity of particles that affect the voxel region at time $t$. And the velocity of a particle is $\ve{v}_p = \frac{\ve{p}_{t+\delta t} - \ve{p}_{t}}{\delta t}$. 

We compare the our dynamic model with TiNeuVox \cite{Fang22_TINeuVox}, a representative method for deformation fields. Specifically, we calculate the mean MFE metric between the motion fields estimated by each model and the corresponding ground truth for sampled time steps $t=[0.1, 0.3, 0.5, 0.7, 0.9]$. To compute \eqref{eq:mfe}, we employ $N=30^3$ voxels evenly distributed within the cubic bounding box of the tested scene. The small time increment $\delta t$ used for velocity approximation is set to $0.01$.
As demonstrated in Tab.~\ref{tab:motion_eval}, particle-based dynamic model in DAP-NeRF effectively captures motion with lower error in dynamic scenes.

\subsection{Method Analysis\label{sec:ablation}}

\begin{figure}[htbp]
    \centering
    \includegraphics[width=0.65\columnwidth]{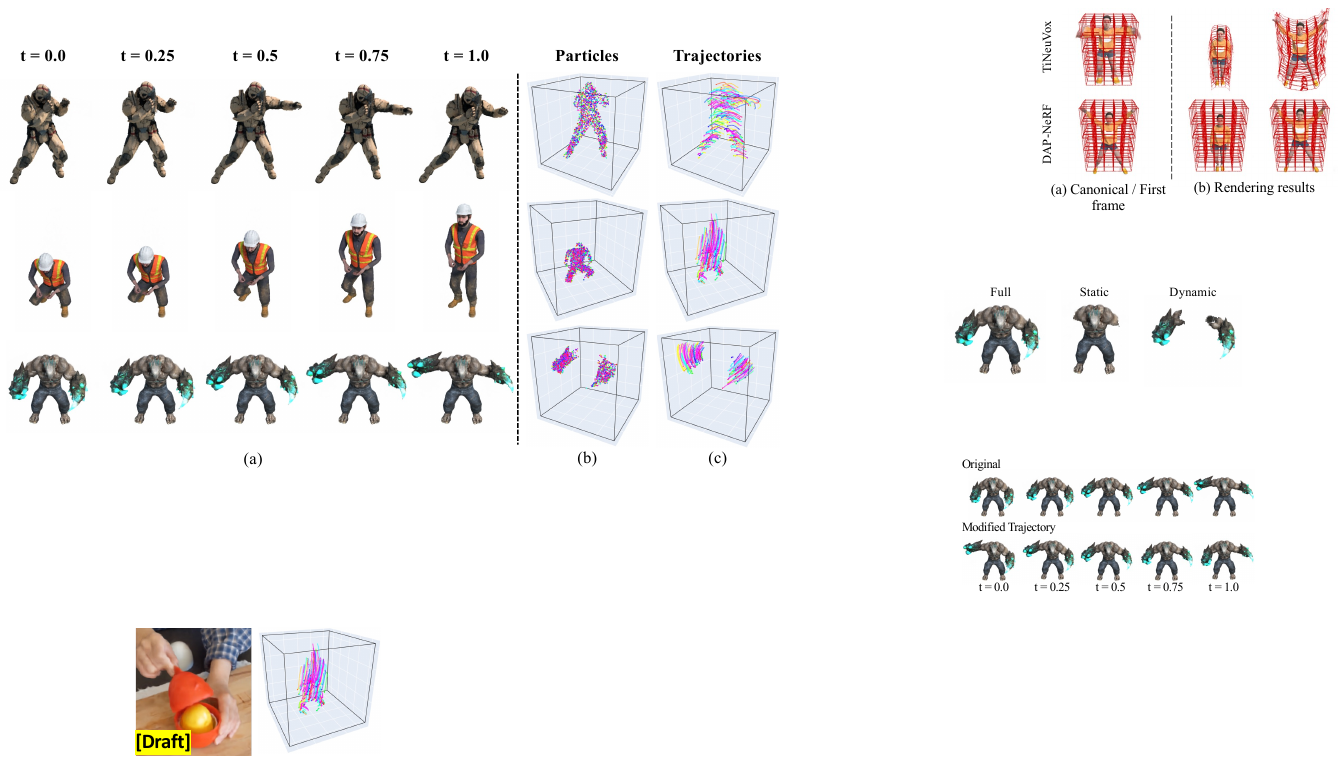}
    \caption{\label{fig:mechanism_comparison} Comparison in dealing with dynamics. We add artificial grids (in red) aligned with the coord-axes at the canonical/first frame for TiNeuVox and our method.}
\end{figure}
\begin{figure}[htbp]
    \centering
    \includegraphics[width=0.65\columnwidth]{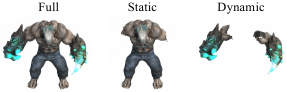}
    \caption{\label{fig:decomposition} Decomposition of scene components. The `full' image is rendered using the final superpositional feature field. The other two images are rendered using only static and dynamic field respectively.}
\end{figure}
\begin{figure}[htbp]
    \centering
    \includegraphics[width=0.8\columnwidth]{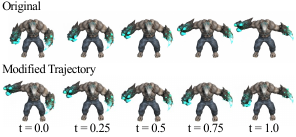}
    \caption{\label{fig:particle_editing} Application on editing dynamic scene. We reverse the original trajectories of particles that are responsible for the shape's right hand.}
\end{figure}
\paragraph{\textbf{Decoupling of Scene Components}}
To illustrate how different models manage dynamics, we introduce artificial grid lines parallel to the axes in the canonical (first) frame, as shown in Fig.~\ref{fig:mechanism_comparison} (a). Note that the grid lines are solely for the visualization of modeled motions; the models are NOT aware of the added lines during training.

The rendering results are shown in Fig.~\ref{fig:mechanism_comparison} (b), which demonstrate that the deforming-based method causes continuous deformation across all space, including empty areas (e.g., air). In comparison, our method decouples the actural moving objects from the static components and only models the motion for objects. Fig.~\ref{fig:decomposition} shows that a trained DAP-NeRF enables a direct and high-quality decomposition. Fig.~\ref{fig:particle_editing} illustrates how this feature can be helpful in practical applications, such as scene editing.

\begin{table}[htbp]
    \centering
    \caption{Ablation study of particle quantity.}
    \begin{tabular}{c|cc}
    \toprule
    \textbf{Particle Number} & \textbf{PSNR$\uparrow$} & \textbf{SSIM$\uparrow$} \\
    \midrule

    $200$k  & \textbf{34.06} & \textbf{0.979} \\
    $100$k  & 33.77 & 0.979 \\
    $50$k   & 33.22 & 0.974 \\
    $10$k   & 32.52 & 0.970 \\
    \bottomrule
    \end{tabular}
    \label{tab:ablation_study}
\end{table}
\begin{figure}[htbp]
    \centering
    \includegraphics[width=0.5\columnwidth]{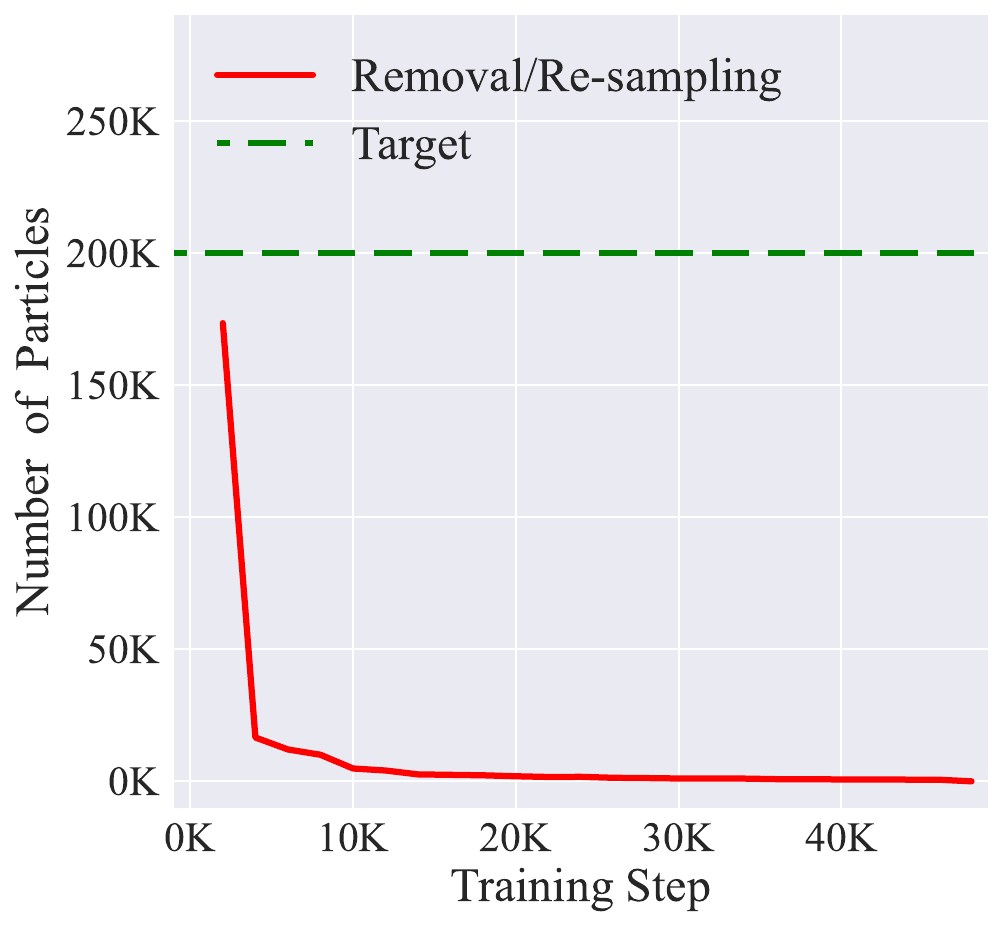}
    \caption{\label{fig:rem_res_curv} This figure demonstrates the effect of particle removal and re-sampling scheme.}
\end{figure}

\paragraph{\textbf{Ablation study of Particle Quantity}}
We examine the impact of particle quantity on performance, employing synthetic scenes from \cite{Pumarola21_D-NeRF} for evaluation. Averaged metrics across all scenes are displayed in Tab. \ref{tab:ablation_study}, which illustrate that the model maintains robust performance even with reduced particle amount.

\begin{table}[htbp]
    \caption{Ablation studies of Loss Components. We conduct evaluations on the D-NeRF dataset \cite{Pumarola21_D-NeRF} and report the average PSNR across all scenes.}
    \centering
    \begin{tabular}{c@{\quad}c@{\quad}c@{\quad}c@{\quad}|@{\quad}c}
    \toprule
    $\mathcal L_\text{ptrgb}$ & $\mathcal L_\text{bg}$ & $\mathcal L_\text{tvf}$ & $\mathcal L_\text{tvm}$ & PSNR$\uparrow$ \\
    \midrule
    $\text{\sffamily x}$ & $\text{\sffamily x}$  & $\text{\sffamily x}$  & $\text{\sffamily x}$  & 33.51 \\
    $\text{\sffamily x}$ & $\text{\sffamily x}$  & $\text{\sffamily x}$  & $\checkmark$          & 33.60 \\
    $\text{\sffamily x}$ & $\text{\sffamily x}$  & $\checkmark$          & $\text{\sffamily x}$  & 33.57 \\
    $\text{\sffamily x}$ & $\text{\sffamily x}$  & $\checkmark$          & $\checkmark$          & 33.62 \\
    $\text{\sffamily x}$ & $\checkmark$          & $\text{\sffamily x}$  & $\text{\sffamily x}$  & 33.72 \\
    $\text{\sffamily x}$ & $\checkmark$          & $\text{\sffamily x}$  & $\checkmark$          & 33.79 \\
    $\text{\sffamily x}$ & $\checkmark$          & $\checkmark$          & $\text{\sffamily x}$  & 33.77 \\
    $\text{\sffamily x}$ & $\checkmark$          & $\checkmark$          & $\checkmark$          & 33.91 \\
    $\checkmark$         & $\text{\sffamily x}$  & $\text{\sffamily x}$  & $\text{\sffamily x}$  & 33.58 \\
    $\checkmark$         & $\text{\sffamily x}$  & $\text{\sffamily x}$  & $\checkmark$          & 33.69 \\
    $\checkmark$         & $\text{\sffamily x}$  & $\checkmark$          & $\text{\sffamily x}$  & 33.67 \\
    $\checkmark$         & $\text{\sffamily x}$  & $\checkmark$          & $\checkmark$          & 33.76 \\
    $\checkmark$         & $\checkmark$          & $\text{\sffamily x}$  & $\text{\sffamily x}$  & 33.79 \\
    $\checkmark$         & $\checkmark$          & $\text{\sffamily x}$  & $\checkmark$          & 33.83 \\
    $\checkmark$         & $\checkmark$          & $\checkmark$          & $\text{\sffamily x}$  & 33.98 \\
    $\checkmark$         & $\checkmark$          & $\checkmark$          & $\checkmark$          & \textbf{34.06} \\
    \bottomrule
    \end{tabular}
    \label{tab:ablation_losses}
\end{table}
\begin{figure}[htbp]
    \centering
    \includegraphics[width=0.6\columnwidth]{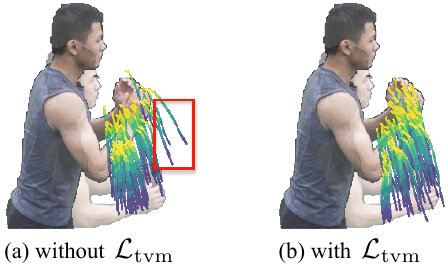}
    \caption{\label{fig:motion_reg_compare} Impact of Motion Regularization Term $\mathcal L_\text{tvm}$. To enhance visual clarity, only 500 particles randomly sampled from around the hand area are rendered.}
\end{figure}
\paragraph{\textbf{Ablation study of Loss Components}} We examine the impact of four auxiliary loss components used in our model, reporting averaged metrics across all scenes from the D-NeRF Dataset \cite{Pumarola21_D-NeRF}. Tab. \ref{tab:ablation_losses} shows that each component contributes to improved rendering quality. 

Specifically, the two terms $\mathcal{L}_\text{ptrgb}$ and $\mathcal{L}_\text{bg}$ help the radiance field better distinguish between scene objects and empty space. The total variation loss $\mathcal{L}_\text{tvf}$ enhances the continuity of the feature field. The combination of these terms effectively prevents overfitting of the model, especially when dealing with monocular videos that have limited multi-view constraints. Additionally, the motion regularization term $\mathcal{L}_\text{tvm}$ encourages particles to preserve local rigidity, leading to more reasonable trajectories and better reconstruction quality. To further validate its effectiveness, we visualized particle trajectories learned with and without $\mathcal{L}_\text{tvm}$. Fig.~\ref{fig:motion_reg_compare} reveals that this motion regularization can effectively reduce artifacts, such as spurious particle trajectories that inaccurately model empty areas.

\paragraph{\textbf{Quantitative Analysis of Removal and Re-sampling}}
Fig. \ref{fig:rem_res_curv} quantitatively demonstrates the impact of alternating removal and re-sampling processes during training. The number of particles re-sampled, which equals the number removed, gradually decreases to zero as the model converges (e.g., after $10$K steps for D-NeRF dataset).

\paragraph{\textbf{Selection of Loss Component Weights}}
The selection of the weights for the four auxiliary losses in \eqref{eq:loss_items} is empirical. Specifically, the weights $w_1$ and $w_2$, corresponding to $\mathcal{L}_\text{ptrgb}$ and $\mathcal{L}_\text{bg}$, follow the common settings used in \cite{Sun22_DVGO, Fang22_TINeuVox}. The other two weights were selected based on hyperparameter tuning, as shown in Tab. \ref{tab:weight_selection}, where $w_3=w_4=0.01$ performed slightly better than smaller values. Additionally, using larger weights would hinder the model's learning.

\begin{table}[htbp]
    \caption{Tuning of weights for $\mathcal L_\mathrm{tvf}$ and $\mathcal L_\mathrm{tvm}$.}
    \centering
    \begin{tabular}{c@{\quad}c@{\quad}|@{\quad}c}
    \toprule
    weight for $\mathcal L_\text{tvf}$ & weight for $\mathcal L_\text{tvm}$ & PSNR$\uparrow$ \\
    \midrule
    0.1   &  0.1   & 32.35 \\
    0.1   &  0.01  & 32.77 \\
    0.1   &  0.001 & 32.73 \\
    0.01  &  0.1   & 33.04 \\
    0.01  &  0.01  & \textbf{34.06} \\
    0.01  &  0.001 & 34.01 \\
    0.001 &  0.1   & 32.89 \\
    0.001 &  0.01  & 33.92 \\
    0.001 &  0.001 & 33.84 \\
    \bottomrule
    \end{tabular}
    \label{tab:weight_selection}
\end{table}

\subsection{Practical Application\label{sec:applications}}
\begin{figure}
    \centering
    \includegraphics[width=0.6\columnwidth]{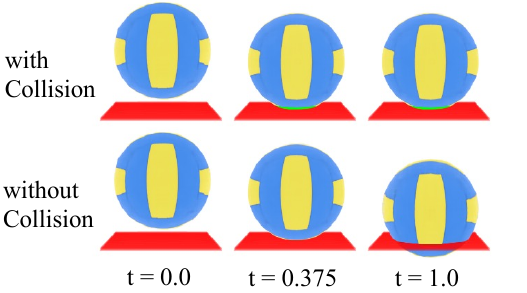}
    \caption{\label{fig:collision} Visualization of collision detection. The red plane represents a virtual wall, while the collided surfaces of the ball are marked in green.}
\end{figure}
We provide an example to present the potential of our models for practical applications. We first train a dynamic NeRF for the scene where a ball slowly falls. When placing a wall in the middle of the ball's path, we can easily detect collision since our model makes an explicitly interpretable and physically meaningful description (i.e., particles) of object motions. As shown in Fig. \ref{fig:collision}, when some particles touch the wall, we can stop the moving object and then mark the collided surfaces of the object by changing the color of the radiance field region affected by those collided particles.
\subsection{Comparison with 3D Gaussian Splatting \label{sec:compare_4DGS}}
\begin{table}[htbp]
    \centering
    \caption{Comparison with 3D Gaussian Splatting on the D-NeRF Dataset \cite{Pumarola21_D-NeRF}. The rendering resolution is set to 800×800.}
    \resizebox{\linewidth}{!}{
    \begin{tabular}{c|l|ccc|c}
    \toprule
    \textbf{Rendering}  &  \textbf{Method} & \textbf{PSNR}$\uparrow$ & \textbf{SSIM}$\uparrow$ & \textbf{LPIPS}$\downarrow$ & \textbf{FPS}$\uparrow$\\
    \midrule
    \multirow{2}{1.2cm}{\textbf{NeRF} \cite{Mildenhall20}}   &TiNeuVox-B \cite{Fang22_TINeuVox} & 32.67  & 0.97  & 0.04 & 1.5 \\
                                                                              &DAP-NeRF (ours)  & 33.84  & 0.98  & 0.02 & 2.7 \\
    \midrule
    \multirow{2}{1.2cm}{\textbf{3DGS} \cite{3DGS}}   &4D-GS \cite{4DGS}       & 34.05  & 0.98  & 0.02 & 82.0 \\
                                         &Deformable-GS \cite{DeformableGS}   & 39.51  & 0.99  & 0.01 & 65.9 \\
    \bottomrule
    \end{tabular}
    }
    \label{tab:nerf_3dgs}
\end{table}

Concurrently, models based on 3D Gaussian Splatting (3DGS) have emerged \cite{DeformableGS, 4DGS}. These models demonstrate strong performance in both speed and quality due to their explicit representation and efficient rasterization \cite{3DGS}. Notably, both 3DGS and our approach utilize explicit primitives to represent small, finite volumes in a scene (Gaussians in 3DGS vs. appearance particles in our method).

Tab. \ref{tab:nerf_3dgs} compares the two branches of methods (NeRF-based and 3DGS-based), showing that 3DGS-based methods achieve impressive results, particularly in terms of speed. In contrast, NeRF-based methods are limited by a more complex rendering process that requires dense evaluation of the neural field.

Nevertheless, the proposed method in this paper benefits from a more regularized motion model, where particles are exclusively responsible for representing movement within the scene, unlike 3DGS-based methods where the entire scene undergoes transformation \cite{DeformableGS, 4DGS}. Furthermore, our method allows the integration of more sophisticated information into the particle features, which can be leveraged for subsequent physics-related tasks.

\section{Conclusion}
We have presented the Dynamic Appearance Particle Neural Radiance Field (DAP-NeRF), a novel framework that introduces Lagrangian particles to construct a superpositional radiance field. The proposed appearance particles can not only carry local light radiance information but also capture object motions in an explicitly interpretable and physically meaningful manner. DAP-NeRF is effective and efficient, requiring only monocular video photometric supervision. We have demonstrated that DAP-NeRF performs well in conventional novel view synthesis and excels in motion modelling tasks. One potential future work is the applications on scenes containing fluids.

\bibliographystyle{plain}
\bibliography{biblo.bib}
\begin{IEEEbiography}
[{\includegraphics[width=1in,height=1.25in,clip,keepaspectratio]{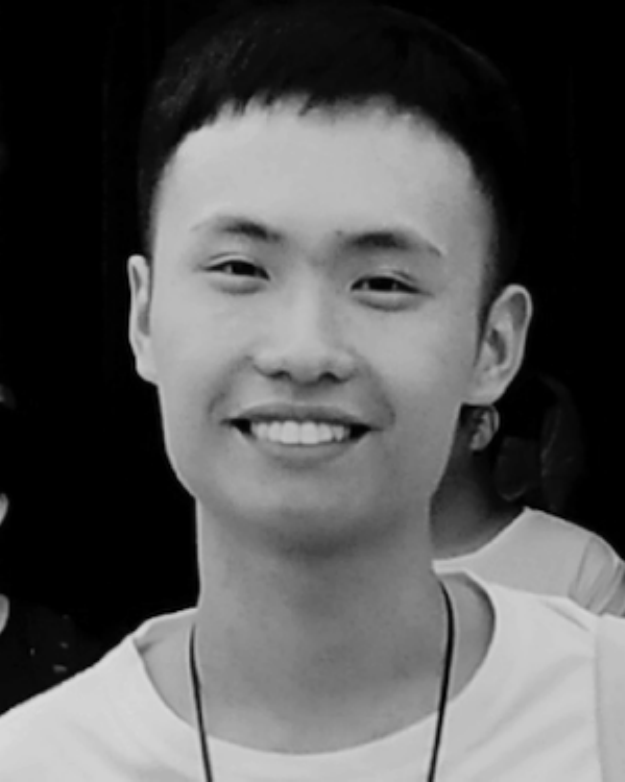}}]{Ancheng Lin}
received the B.Eng. degree in computer science from Guangdong Polytechnic Normal University, Guangdong, China, in 2019. He is currently pursuing the Ph.D. degree with the Faculty of Engineering and Information Technology, School of Computer Science, University of Technology Sydney, Australia. His research interests include 3D vision, autonomous driving, and deep learning.

\end{IEEEbiography}

\begin{IEEEbiography}
[{\includegraphics[width=1in,height=1.25in,clip,keepaspectratio]{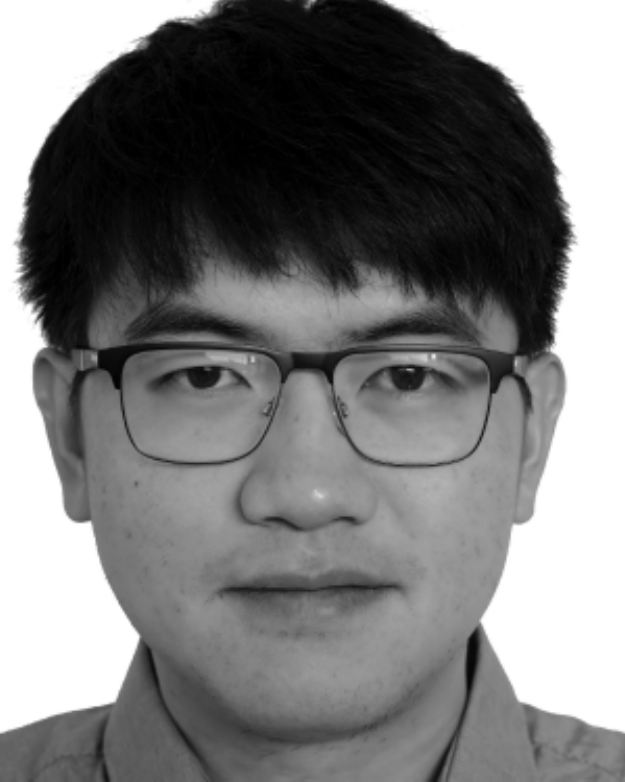}}]{Yusheng Xiang} (Member, IEEE) received the M.Sc. degree in vehicle engineering from the Karlsruhe Institute of Technology, Karlsruhe, Germany, in 2017, with a focus on mathematical model building and simulation, where he also got the Ph.D. degree from the Institute of Vehicle System Technology in 2021. He was a Research Scientist with Robert Bosch GmbH, Germany. From September 2020 to February 2021, he was a Visiting Scholar at the University of California at Berkeley, USA, supervised by Prof. Samuel S. Mao. He is also the CTO of Elephant Tech LLC, Shenzhen, China, a spinoff from Prof. Samuel S. Mao’s Laboratory. He is a Lecturer with the School of Automobile, Chang'an University, Xi'an, China. He has authored nine influential journals and international conference papers, and holds more than twenty patents. His group deals with improving mobile machines’ performance using artificial intelligence and the Internet of Things. 
\end{IEEEbiography}

\begin{IEEEbiography}
[{\includegraphics[width=1in,height=1.25in,clip,keepaspectratio]{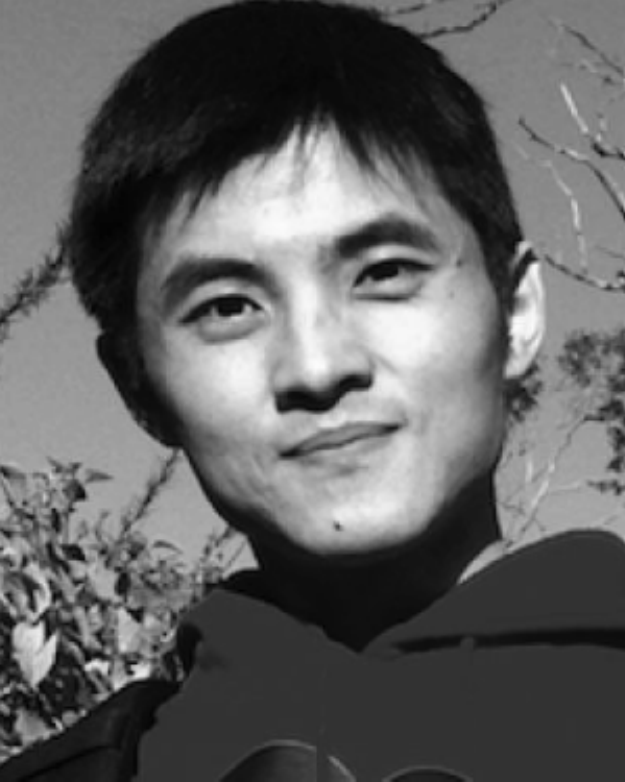}}]{Jun Li} received the B.S. degree in computer science and technology from Shandong University, China, in 2003, the M.Sc. degree in information and signal processing from Peking University, Beijing, China, in 2006, and the Ph.D. degree in computer science from Queen Mary University of London, U.K., in 2009. He is currently a Senior Lecturer with the Artificial Intelligence Institute, and the Faculty of Engineering and Information Technology, School of Computer Science, University of Technology Sydney, Australia. He also leads the AI Department, Elephant Tech LLC. His research interests include probabilistic data models and image and video analysis using neural networks.
\end{IEEEbiography}

\begin{IEEEbiography}
[{\includegraphics[width=1in,height=1.25in,clip,keepaspectratio]{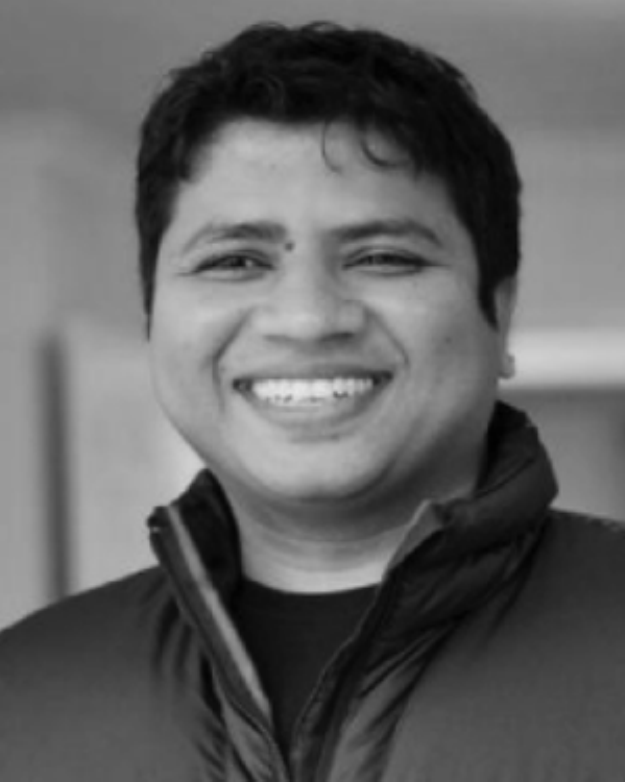}}]{Mukesh Prasad}
(Senior Member, IEEE) is a Senior Lecturer at the School of Computer Science in the Faculty of Engineering and IT at UTS who has made substantial contributions to the fields of machine learning, artificial intelligence and Natural Language Processing.  He is working also in the evolving and increasingly important field of image processing, data analytics and edge computing, which promise to pave the way for the evolution of new applications and services in the areas of healthcare, biomedical, agriculture, smart cities, education, marketing and finance. His research has appeared in numerous prestigious journals, including IEEE/ACM Transactions, and at conferences, and he has written more than 150 research papers. He started his academic career as a lecturer with UTS in 2017 and became a core member of the University’s world-leading Australian Artificial Intelligence Institute (AAII), which has a vision to develop theoretical foundations and advanced technologies for AI and to drive progress in related areas. His research is backed by industry experience, specifically in Taiwan, where he was the principal engineer (2016-17) at the Taiwan Semiconductor Manufacturing Company (TSMC). There, he developed new algorithms for image processing and pattern recognition using machine learning techniques. He received an M.S. degree from the School of Computer and Systems Sciences at the Jawaharlal Nehru University in New Delhi, India (2009), and a PhD from the Department of Computer Science at the National Chiao Tung University in Taiwan (2015).
\end{IEEEbiography}

\end{document}